%% file: main.tex
\documentclass{article}

\usepackage{microtype}
\usepackage{graphicx}
\usepackage{subfigure}
\usepackage{booktabs} %
\usepackage[ruled, vlined]{algorithm2e}    %

\usepackage{xcolor}
\usepackage{hyperref}
\usepackage[textwidth=3cm]{todonotes}
\usepackage{rotating}
\usepackage{amssymb}
\usepackage{amsthm}
\newtheorem{theorem}{Theorem}[section]

\newtheorem{lemma}[theorem]{Lemma}
\usepackage[accepted]{icml2020}

\input{math_commands.tex}

\newcommand{\htr}{\hat{\vh}_{\vtheta,\vgamma}}
\newcommand{\Jtr}{\hat{J}_{\vtheta,\vgamma}}
\newcommand{\Dtr}{\hat{\Delta}_{\vtheta_t,\vgamma}}
\newcommand{\ptheta}{p_\vtheta}
\newcommand{\pthetat}{p_{\vtheta_t}}
\newcommand{\qphi}{{q_\vphi}}
\newcommand{\Dx}{{\Delta_\vtheta(\vx)}}
\newcommand{\nat}{\veta}
\newcommand{\natt}{{\nat_\vtheta}}
\newcommand{\suff}{{\vs}}
\newcommand{\lnorm}{{\Psi}}
\newcommand{\lnormt}{\lnorm_\vtheta}
\newcommand{\pjoint}{\ptheta(\vz, \vx)}
\newcommand{\lpjoint}{\log \ptheta(\vz, \vx)}
\newcommand{\nlpjoint}{\ntheta \log \ptheta(\vz, \vx)}
\newcommand{\lptheta}{\log \ptheta}
\newcommand{\Jtheta}{J_{\vtheta}}
\newcommand{\nJtheta}{\ntheta \Jtheta}
\newcommand{\nlptheta}{\ntheta \log \ptheta}
\newcommand{\lpmargin}{\lptheta(\vx)}

\newcommand{\ppost}{\ptheta(\vz|\vx)}
\newcommand{\ppostt}{\pthetat(\vz|\vx)}
\newcommand{\evalt}{\big\rvert_{\vtheta_t}}
\newcommand{\Dxt}{{\Delta_{\vtheta_t}(\vx)}}
\renewcommand{\E}[2]{\mathbb{E}_{#1}{\left[#2\right]}}
\newcommand{\ntheta}{{\nabla_\vtheta}}
\newcommand\numberthis{\addtocounter{equation}{1}\tag{\theequation}}

\newcommand{\httpsurl}[1]{\href{https://#1}{\nolinkurl{#1}}}

\usepackage[capitalize,noabbrev]{cleveref}

\newcommand{\mytitle}{Amortised Learning by Wake-Sleep}

\icmltitlerunning{\mytitle}

\begin{document}

\twocolumn[
\icmltitle{\mytitle}

\icmlsetsymbol{equal}{*}

\begin{icmlauthorlist}
\icmlauthor{Li K. Wenliang}{gatsby}
\icmlauthor{Theodore Moskovitz}{gatsby}
\icmlauthor{Heishiro Kanagawa}{gatsby}
\icmlauthor{Maneesh Sahani}{gatsby}
\end{icmlauthorlist}

\icmlaffiliation{gatsby}{Gatsby Computational Neuroscience Unit}

\icmlcorrespondingauthor{Li K. Wenliang}{kevinli@gatsby.ucl.ac.uk}
\icmlkeywords{Machine Learning, ICML}

\vskip 0.3in
]

\printAffiliationsAndNotice{}  %

\begin{abstract}
Models that employ latent variables to capture structure in observed
data lie at the heart of many current unsupervised learning
algorithms, but exact maximum-likelihood learning for powerful and flexible
latent-variable models is almost always intractable. Thus, state-of-the-art
approaches either abandon the maximum-likelihood framework entirely, or
else rely on a variety of variational approximations to the
posterior distribution over the latents.
Here, we propose an alternative approach that we call amortised learning.
Rather than computing an approximation to the posterior over
latents, we use a wake-sleep Monte-Carlo strategy to learn a function
that directly estimates the maximum-likelihood parameter updates.
Amortised learning is possible whenever samples of latents
and observations can be simulated from the generative model, treating the model
as a ``black box''.
We demonstrate its effectiveness on a wide range of complex models,
including those with latents that are discrete or supported on non-Euclidean spaces.
\end{abstract}

\section{Introduction} \label{sec:introduction}

Many problems in machine learning, particularly unsupervised learning,
can be approached by fitting flexible parametric probabilistic models
to data, often based on ``local'' latent variables whose number
scales with the number of observations.
Once the optimal parameters are found, the resulting model may be used
to synthesise samples, detect outliers, or relate observations to a
latent ``representation''.
The quality of all of these operations depends on the appropriateness
of the model class chosen and the optimality of the identified
parameters.

Although many fitting objectives have been explored in the literature,
maximum-likelihood (ML) estimation remains prominent and comes with attractive theoretical
properties, including consistency and asymptotic efficiency
\citep{NeweyMcFadden1994Large}.
A challenge, however, is that analytic evaluation of the likelihoods of rich, 
flexible latent variable models is usually intractable.
The Expectation-Maximisation (EM) algorithm
\citep{DempsterRubin1977Maximum} offers one route to ML estimation in
such circumstances, but it in turn requires an explicit calculation of
(expected values under) the posterior distribution over latent variables, 
which also proves to be intractable in most cases of interest.
Consequently, state-of-the-art ML-related methods almost always rely on
approximations, particularly in large-data settings.

Denote the joint distribution of a generative model as $\ptheta(\vz,\vx)$ where $\vz$ is latent and $\vx$ is observed, and $\vtheta$ is the vector of parameters. 
EM breaks the ML problem into an iteration of 
two sub-problems.  Given parameters $\vtheta_t$ on the $t$th iteration,
first find the posterior $\pthetat(\vz|\vx)$; then maximise a lower 
bound to the likelihood that depends on this posterior to 
obtain $\vtheta_{t+1}$.  This bound is tight when computed using the correct
posterior, ensuring convergence to a local mode of the likelihood.

The intractability of $\ptheta(\vz|\vx)$ forces some combination of 
Monte-Carlo estimation and the use of a tractable parametric 
approximating family which we call $q(\vz|\vx)$ 
\citep{BishopBishop2006Pattern}.  
To avoid repeating the expensive optimisation in finding $q(\vz|\vx)$ for each $\vx$,
amortised inference trains an encoding or recognition model, with parameters $\vphi$, to 
map from any $\vx$ directly to an approximate posterior $\qphi(\vz|\vx)$. 
Examples of amortised inference models include the Helmholtz machine
\citep{dayan1995helmholtz, hinton1995wake} trained by the wake-sleep algorithm;
and the variational auto-encoder (VAE) \citep{KingmaWelling2014,RezendeEtAl2014} trained using
reparamerisation gradient methods.
With considerable effort on improving variational inference (reviewed in \citep{zhang2018advances}),
complex and flexible generative models have been trained on large, high-dimensional datasets.

\begin{figure}[!ht]
    \centering
    \includegraphics[width=0.7\columnwidth]{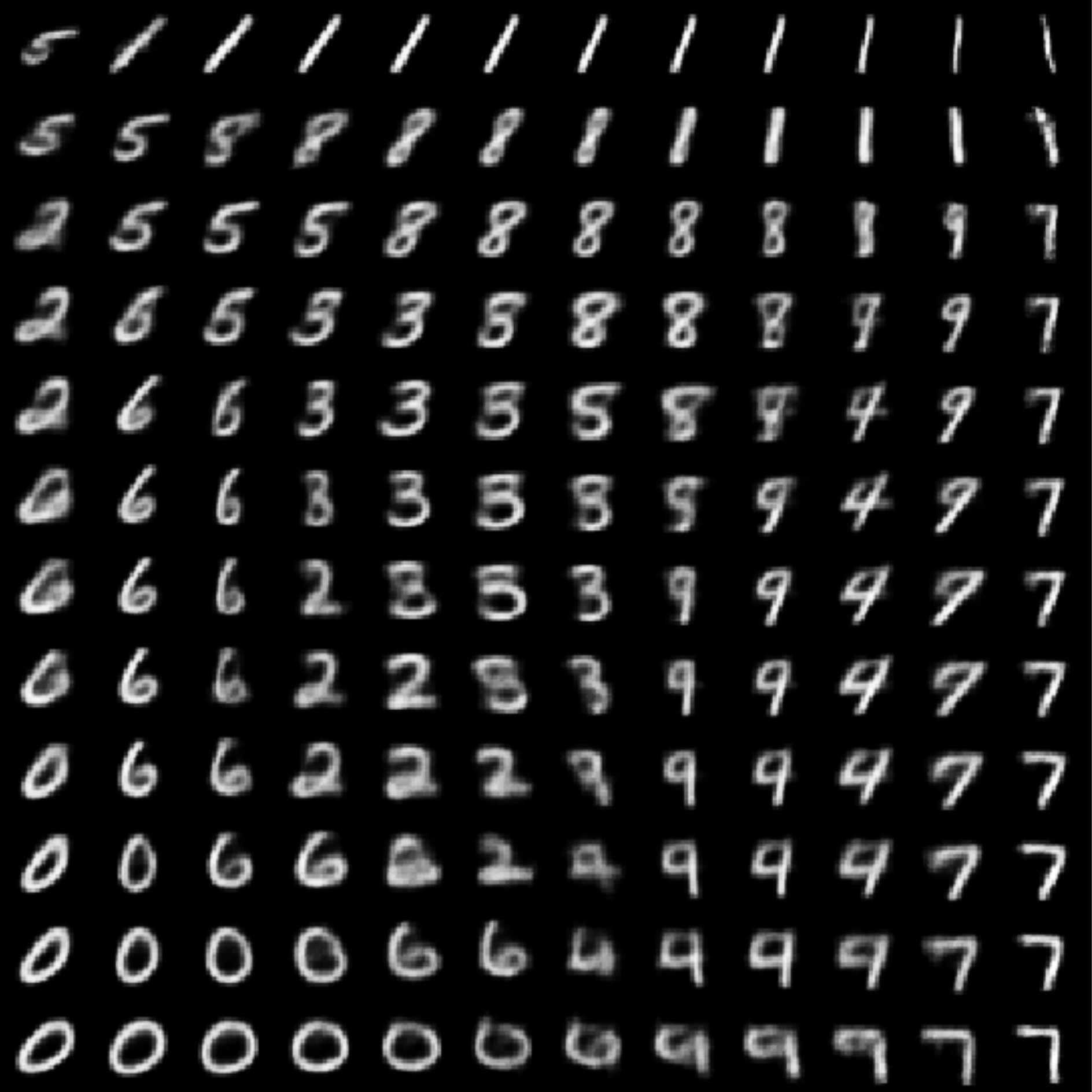}
    \includegraphics[width=0.9\columnwidth]{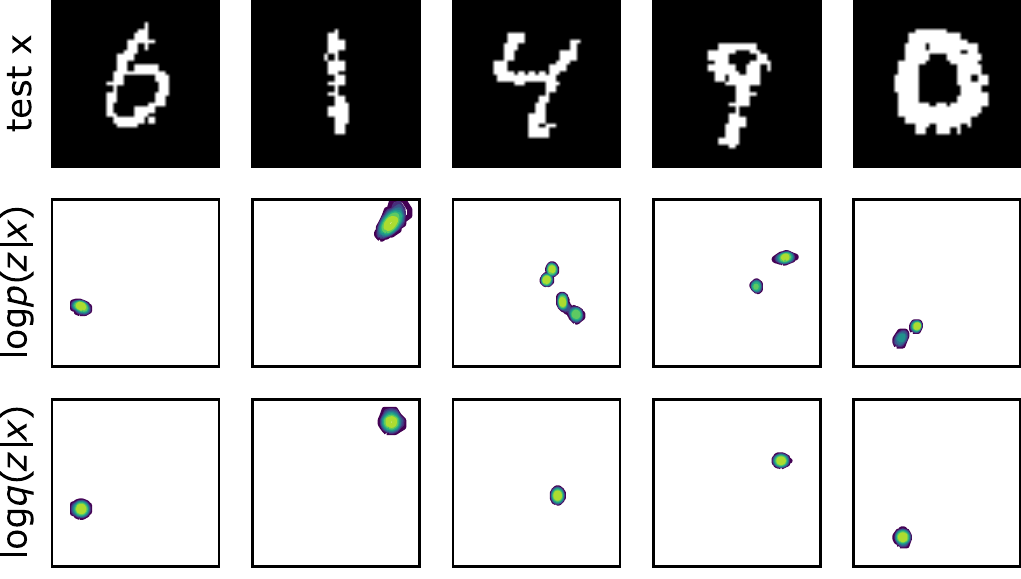}
    \caption{VAE trained on binarised MNIST digits. 
    Top: mean images generated by decoding points on a grid of 2-D latent variables. 
    Bottom three rows show five samples of real MNSIT digit (top), 
    the corresponding true posteriors (middle) found by histogram and 
    the approximate posteriors computed by the encoder.}
    \label{fig:bimodal}
\end{figure}

However, approximate variational inference poses at least three challenges.
First, the parametric form of the approximate posterior $q(\vz|\vx)$, and particularly any 
factorisations assumed, must be crafted for each model.  
Second, methods such as reparameterisation require specific transformations tailored to the
type of latent variables, whether they are continuous or discrete, and whether or not the support is Euclidean.
Third, given a flexible generative model, such as one with conditional dependence 
modelled using neural networks, the true posteriors may be irregular in ways that are difficult 
to approximate.
We illustrate this latter effect using a standard VAE with two-dimensional $\vz$ trained on
binarised MNIST digits (\cref{fig:bimodal}).
The exact posterior may be distorted or multi-modal, even though only 
Gaussian posteriors are ever produced by the encoder.

When inference is only approximate, the M-step of EM may not increase 
the likelihood, and so approximate methods usually converge away from the ML 
parameter values.  The dependence of learnt parameters on the quality of the 
posterior approximation is not straightforward, and the error may not be reduced by
(say) approximations with lower Kullback-Leibler (KL) divergence \citep{TurnerSahani2011Two}; 
indeed errors in posterior statistics that enter the objective function may be unbounded 
\citep{HugginsBroderick2019Practical}.

Here, we propose a novel approach to ML learning in flexible latent variable models
that avoids the complications of posterior estimation, 
instead learning to predict the gradient of 
the likelihood directly---an approach we call \emph{amortised learning}.
The particular realisation we develop here, amortised learning by wake sleep (ALWS), 
requires only that sampling from the generative model $\ptheta(\vz,\vx)$ be possible, 
and that the gradient $\nlpjoint$ be available (possibly by automated methods), 
but otherwise does not make assumptions about the latent variable form or distribution.
We test the performance of ALWS on a wide range of tasks and models,
including hierarchical models with heterogeneous priors, 
nonlinear dynamical systems, 
and deep models of images.
All experiments use the same form of gradient model 
trained by simple least-squares regression.
For image generation, we find that models trained with ALWS can produce samples of
considerably better quality than those trained using algorithms based on 
variational inference.

\section{Background} \label{sec:background}

\subsection{Model Definition}
Consider a probabilistic generative model  with parameter vector $\vtheta$ that defines a prior on latents 
$p_\vtheta(\vz)$ and a conditional on observations $p_\vtheta(\vx|\vz)$.
In ML learning, we seek parameters that maximise the log (marginal) likelihood 
\begin{equation}\label{eq:loglik}
    \log\ptheta(\vx)=\log \int \ptheta(\vz)\ptheta(\vx|\vz) \ud\vz
\end{equation}
averaged over a set of i.i.d.\ data $\gD=\{\vx_m^*\}_{m=1}^M$. 
One approach is to iteratively update $\vtheta$ by following the gradient
\begin{gather}\label{eq:loglik_grad}
    \Dx := \ntheta \lpmargin
\end{gather}
at each iteration\footnote{We define the likelihood gradient for a single data point here and throughout; an actual update will typically follow the gradient averaged over i.i.d\  data.}

\subsection{Variational Inference for Learning}\label{sec:vi}

For many models of interest, the integral in \eqref{eq:loglik} cannot be evaluated analytically, and 
so direct computation of the gradient is intractable. 
A popular alternative is to maximise a variational
    lower bound on the marginal likelihood defined by a distribution $q(\vz)$:
\begin{equation}\label{eq:free_energy}
    \gF(q, \vtheta) := \E{q(\vz)}{\log \ptheta(\vz,\vx)} + \sH[q]\le \log \ptheta(\vx),
\end{equation}
where $\sH[q]$ is the entropy of $q$. 
Thus, the parameter $\vtheta$ can be updated by following the gradient of $\gF(q,\vtheta)$ w.r.t.\ $\vtheta$
\begin{align*}
    \ntheta \gF(q, \vtheta) &= \ntheta\E{q(\vz)}{\lpjoint} \\
                            &= \E{q(\vz)}{\ntheta \lpjoint}. \numberthis \label{eq:free_energy_grad}
\end{align*}
When $q(\vz)=\ptheta(\vz|\vx)$, the lower bound in \eqref{eq:free_energy} is tight, 
and the gradient in \eqref{eq:free_energy_grad} is equal to that of the likelihood 
(see \cref{sec:Dx_proof}).
Variational approximations attempt to bring $q$ close to $\ptheta(\vz|\vx)$, usually by seeking to minimise $\KL[q(\vz)||\ptheta(\vz|\vx)]$ (which corresponds to maximising the bound $\gF$ w.r.t.\ $q$). 
However, although minimising $\KL[q(\vz)||\ptheta(\vz|\vx)]$ over $q$ ensures consistent 
optimisation of a single objective, the resulting gradient in \eqref{eq:free_energy_grad} will 
often be a poor approximation to the likelihood gradient \eqref{eq:loglik_grad}.

\subsection{Conditional Expectation and LSR}\label{sec:cond_exp_lsr}

Our approach is to avoid the difficulties introduced by approximating $\ptheta(\vz|\vx)$ with $q(\vz)$ in 
\eqref{eq:free_energy_grad}, and instead estimate the conditional expectation 
directly using least-squares regression (LSR). 
Let $\vx$ and $\vy$ be random vectors with a joint distribution $\rho(\vx,\vy)$ on $\sR^{d_x}\times \sR^{d_y}$. 
In LSR, we seek a (vector-valued) function $\vf$ that achieves the lowest 
mean squared error (MSE) $\E{\rho(\vx,\vy)}{\lVert \vy - \vf(\vx)\rVert^2_2}$. 
The ideal solution is given by 
$\vf_\rho(\vx):=\E{\rho(\vy|\vx)}{\vy}$, as   
the problem can be cast as the minimisation of  $\E{\rho(\vx)}{\|\vf_{\rho}(\vx)-\vf(\vx)\|_2^2}$, where $\rho(\vx)$ is the marginal distribution of $\vx$ (see \cref{sec:mse_for_mean_proof}).
Note that $\vf_\rho(\vx)$ takes a similar form as the desired \eqref{eq:free_energy_grad}.
In practice, the distribution $\rho(\vx,\vy)$ is known only through a sample $\{(\vx_n,\vy_n)\}_{n=1}^N\stackrel{\text{i.i.d.}}{\sim}\rho(\vx,\vy)$; thus, LSR
can be understood to seek a good approximation of $\vf_{\rho}$ based on the sample. 

\subsection{Kernel Ridge Regression}\label{sec:KRR}
In LSR, as the target $\vf_\rho$ is unknown, it is desirable to construct an estimate without imposing restrictions on its form. 
Kernel ridge regression (KRR) is a nonlinear regression method that draws the estimated regression function 
from a flexible class of functions called a
reproducing-kernel Hilbert space (RKHS) \citep{HofmannSmola2008Kernel}.
The KRR estimator is found by minimising the regularised empirical risk
\begin{equation}\label{eq:krr_loss}
    \min_{\vf\in \gH}\frac{1}{N}\sum_{n=1}^N\|\vy_n - \vf(\vx_n) \|^2_2 +\lambda \|\vf\|^2_\gH,
\end{equation}
where $\lambda>0$ is a regularisation parameter, and $\gH$ is the RKHS corresponding to a matrix-valued kernel $\kappa:\sR^{d_x}\times \sR^{d_x}\to \sR^{d_y\times d_y}$ \citep{CarmeliToigo2006Vector}. 
The solution can be found conveniently in closed-form, which 
allows a further simplification detailed in \cref{sec:krr_wake_sleep}.
In this paper, we use a kernel of the form $\kappa(x,x') = k(x,x')\mI_y$, where $\mI_{y}$ is the identity matrix, and $k$ is a scalar-valued positive definite kernel; therefore, the matrix-valued kernel $\kappa$ can be identified with its scalar counterpart $k$. 
In particular, in the scalar output case $d_y=1$, this choice of $\kappa$ coincides with KRR with 
the scalar kernel $k$. 
Importantly, the closed-form solution $\hat{\vf}_\lambda$ of KRR in \eqref{eq:krr_loss} can be expressed as 
\begin{equation}\label{eq:KRR_prediction}
    \hat{\vf}_{\lambda}(\vx^*) = \mathbf{Y}(\mK+N\lambda \mI_N)^{-1}\vk^*, 
\end{equation}
where $ \mathbf{Y}$ is the concatenation of the training targets $[\vy_1, \dots, \vy_N] \in \sR^{d_y \times N}$,
$\mK \in \sR^{N\times N}$ is the gram matrix whose element is $(\mK)_{ij} = k(\vx_i, \vx_j)$,  
$\mI_N$ is the identity matrix and $\vk^*=(\evk(\vx_i, \vx^*))_{i=1}^N\in \sR^N$ for a test point $\vx^*$.

In the limit of $N\to \infty$ and $\lambda \to 0$, the solution $\hat{\vf}_{\lambda}$ will achieve the minimum MSE in the RKHS \citep{CaponnettoDeVito2007Optimal}. 
In general, the target $\vf_\vrho$ may not be in the RKHS\footnote{In this case, $\vf_\vrho$ is only assumed to be square-integrable with respect to $\vrho$}; 
nonetheless, if the RKHS is sufficiently rich (or $C_0$ universal \citep{CarmeliUmanita2010Vector}), the error made by the estimator  $\E{\rho(\vx)}{\|\hat{\vf}_{\lambda}(\vx)-\vf_{\rho}(\vx)\|_2^2}$ will converge to zero \citep[Theorem 7]{SzaboGretton2016Learning}.

\section{Amortised Learning by Wake-Sleep}\label{sec:Alws}

\subsection{Gradient of Log-Likelihood}
As stated above and derived in \cref{sec:Dx_proof}, the log-likelihood gradient function evaluated on observation $\vx$ at iteration $t$ (with current parameters $\vtheta_t$) can be written
\begin{align*}
 \Dxt   &= \ntheta\log\pthetat(\vx)\evalt \\
        &= \ntheta \gF(\pthetat(\vz|\vx),\vtheta)\evalt, 
        \numberthis \label{eq:free_energy_exact_grad}
\end{align*}
where the gradient in the second line is taken w.r.t.~the second argument 
of $\gF$; 
the posterior distribution is for a fixed $\vtheta$ at the current $\vtheta_t$.

We want to directly estimate of this gradient without explicit computation of the posterior.
Inserting the definition from  \eqref{eq:free_energy_grad} into \eqref{eq:free_energy_exact_grad} we have,
\begin{align*}
    \Dxt &= \E{\pthetat(\vz|\vx)}{\ntheta\lptheta(\vz,\vx)\evalt } \numberthis \label{eq:exact_grad_manual}\\
              &= \ntheta  \E{\pthetat(\vz|\vx)}{\lptheta(\vz,\vx)} \evalt \\
              &= \ntheta  \Jtheta(\vx) \evalt. \numberthis \label{eq:exact_grad_autodiff}
\end{align*}
where $\Jtheta(\vx):=\E{\pthetat(\vz|\vx)}{\lptheta(\vz,\vx)}$. 
Note that the function $\Jtheta(\vx)$ changes with iteration due to the dependence 
on $\pthetat(\vz|\vx)$. It can be regarded 
as an instantaneous objective for ML learning starting from $\vtheta_t$.
Neither \eqref{eq:exact_grad_manual} nor \eqref{eq:exact_grad_autodiff} can be computed in closed form, and therefore need to be estimated. 
We refer to ML learning via the estimation of $\Dxt$ either through $\Jtheta$ by \eqref{eq:exact_grad_autodiff} 
or directly by \eqref{eq:exact_grad_manual} as amortised learning. 
The difference between the two equations  
lies purely in implementation: The former estimates the high-dimensional $\Dxt$ directly, 
whereas the latter implements the same computation by differentiating $\Jtheta(\vx)$. 
We term an estimator of $\Jtheta$ a \textit{gradient model}, as it retains information about $\vtheta$ 
and is used to estimate the gradient $\Dxt$. 
In the next section, we develop a concrete instantiation of amortised learning. 

\subsection{Training KRR Gradient Model by Wake-Sleep}\label{sec:krr_wake_sleep}
As discussed in \cref{sec:cond_exp_lsr}, LSR allows us to estimate the conditional expectation of an 
output variable given an input. 
Thus, although the gradient in \eqref{eq:exact_grad_manual} (or in \eqref{eq:exact_grad_autodiff})
involves an intractable conditional expectation, we can obtain an estimate of the gradient $\Dxt$ by regressing from $\vx$ to $\ntheta\lptheta(\vz,\vx)$ 
(or $\lptheta(\vz,\vx$)). 
Any reasonable regression model, e.g., a neural network, could serve this purpose, 
but here we choose to use KRR introduced in \cref{sec:KRR}. 
Other possible forms of gradient model are discussed in \cref{sec:other_gms}.

The expression in \eqref{eq:exact_grad_manual} leads to the following LSR problem
\begin{equation}\label{eq:manual_loss}
    \min_{\vf\in\tilde{\gH}} \frac{1}{N}\sum_{n=1}^N{\|\ntheta (y_{\vtheta,n})\evalt - \vf(\vx_n) \|_2^2} + \lambda \|\vf\|_{\tilde{\gH}}^2,
\end{equation}
where $y_{\vtheta,n}= \lptheta(\vz_n,\vx_n)$, $\tilde{\gH}$ is an RKHS and $\{(\vz_n,\vx_n)\}_{n=1}^N\sim \pthetat$. 
\citet{BrehmerCranmer2020Mining} also noticed that 
log-likelihood gradient could be obtained by LSR.
However, regressing to a vector-valued $\nlptheta$ can be expensive,
and evaluating the target $y_{\vtheta,n}$ on all $(\vz_n,\vx_n)$ is slow. 
Alternatively, we can use \eqref{eq:exact_grad_autodiff} and find an estimator for the 
scalar-valued $\Jtheta$ that keeps the dependence on $\vtheta$ and then evaluate its gradient
by automatic differentiation.
Thus, we construct an estimator by
\begin{equation}\label{eq:autodiff_loss}
    \Jtr = \argmin_{f\in\gH} \frac{1}{N} \sum_{n=1}^N{|y_{\vtheta,n} - f(\vx_n) |^2} + \lambda \|f\|_\gH^2, 
\end{equation}
where $\gH$ is the RKHS induced by a kernel $k_\vomega(\cdot,\cdot)$ with hyperparameters $\vomega$, and $\vgamma=\{\vomega,\lambda\}$. 
For each data point $\vx^*\in\gD$, the estimate of $\Jtheta(\vx^*)$ is 
\begin{gather*}
    \Jtr(\vx^*) = \valpha_{\vtheta,\vgamma}\cdot\vk_{\vomega}^*,\numberthis \label{eq:krr_J}\\
    \valpha_{\vtheta,\vgamma}=\vy_\vtheta \left(\mK_\vomega + \lambda N\mI_N\right)^{-1},~~
    (\vy_{\vtheta})_n=\lptheta(\vz_n,\vx_n) \\
    \emK_{\vomega,i,j} = k_{\vomega}(\vx_i, \vx_j), \quad  \evk^*_{\vomega,j}=k_\vomega(\vx_j,\vx^*)
\end{gather*}
where $\mI_N$ is the identity matrix of size $N\times N$. 
Note that the dependence of $\Jtr$ on $\vtheta$ is only through 
evaluations of $\log \pjoint$ on samples drawn from $\pthetat$
for fixed $\vtheta=\vtheta_t$.
The gradient $\Dxt$ is then estimated as 
$$
\Dtr(\vx):=\ntheta\Jtr(\vx)\evalt.
$$
In general, a good estimator of $\Jtheta$ may not yield a 
reliable estimate of its gradient $\nJtheta$;
however, for the KRR estimate, taking the derivative of $\Jtr$ w.r.t.\ $\vtheta$ is equivalent to 
replacing $\vy_\vtheta$ in $\eqref{eq:krr_J}$ with $\ntheta (\vy_\vtheta)|_{\vtheta_t}$,
which is the solution for the optimisation in \eqref{eq:manual_loss}, 
with $\tilde{\gH}$ being a vector-valued RKHS given by a kernel $\kappa_{\vomega}=k_{\vomega}\mI$ (see \cref{sec:KRR}). 
We show in \cref{sec:l2p} that,  under mild conditions, 
the target of the regression
$\E{\ppostt}{\ntheta y_{\vtheta,n}\evalt}$ is square-integrable under $\pthetat(\vx)$ for common generative models.

In summary, learning proceeds according to the following wake-sleep procedure: 
at the $t$th step when $\vtheta=\vtheta_t$,
the gradient model is first trained 
using ``sleep samples'' $(\vz_n, \vx_n) \sim \pthetat$ and evaluations $\lptheta(\vz_n,\vx_n)$, 
keeping the dependence on $\vtheta$; 
then the gradient model is applied to
real data (``wake'' samples) $\vx^*\in\gD$ to produce $\Dtr(\vx^*)$ 
by differentiating $\Jtr$ and evaluating at $\vtheta_t$. 
See \cref{alg}.
Two points are worth emphasis:
(a) The algorithm does not require explicit computation or approximation of the posterior, and  
(b) We only need samples from the model $\ptheta(\vz,\vx)$ and differentiable evaluations of $\lpjoint$. 

\subsection{Exponential Family Conditionals}\label{sec:exp_fam_likelihood}
In many common models, the conditional $\ptheta(\vx|\vz)$ lies in the exponential family 
(e.g. Gaussian, Bernoulli), 
and we can exploit this structure to simplify the estimation of $\Jtheta$. 
In this case, the log joint can be written as 
\begin{align*}
    \lpjoint &= \log \ptheta(\vx|\vz) + \log \ptheta(\vz)  \\
             &= \natt(\vz)\cdot \suff(\vx) - \log Z_\vtheta(\vz) + \log \ptheta(\vz) \\
             &= \natt(\vz)\cdot \suff(\vx) - \lnormt(\vz)
\end{align*}
where $\natt(\vz)$, $\suff(\vx)$ and $Z_\vtheta(\vz)$ are, respectively, the natural parameter,
sufficient statistics and normaliser of the likelihood, 
and $\lnormt:=\log Z_\vtheta(\vz) - \log p_\vtheta(\vz)$.
By taking the posterior expectation, 
$J_\vtheta(\vx)$ in \eqref{eq:exact_grad_autodiff} becomes
\begin{align*}
    J_\vtheta(\vx) & =  \underbrace{\E{\pthetat}{\natt(\vz)}}_{\vh_\vtheta^{\nat}(\vx)}\cdot \suff(\vx)- 
                        \underbrace{\E{\pthetat}{\lnormt(\vz)}}_{{h_\vtheta^{\lnorm}(\vx)}} \numberthis \label{eq:expfam_autodiff_grad}
\end{align*}
where $\pthetat$ stands for $\pthetat(\vz|\vx)$.
Therefore, for exponential family likelihoods, 
the regression to $\lpjoint$ in \eqref{eq:autodiff_loss} 
can be replaced by two separate regressions to $\natt(\vz)$ and $\lnormt(\vz)$, 
which are functions of $\vz$ alone. 
The resulting estimators $\htr^\nat$ and $\htr^\lnorm$ are combined to yield 
$$
\Dtr(\vx)=\ntheta\left.\left[\htr^\nat(\vx) \cdot \suff(\vx)\right]\right\rvert_{\vtheta_t}-
\ntheta\htr^\lnorm(\vx)\vert_{\vtheta_t},
$$
where the Jacobian vector product applies to the first term.

\begin{algorithm}[tb]
\SetAlgoLined
\SetKwInOut{Input}{input}
\SetKwInOut{Output}{return}
\Input{Dataset $\gD$, 
gradient model parameters $\vgamma$,
generative model $\lpjoint$, or $\natt$ and $\lnormt$ with 
parameters $\vtheta$ 
initialised s.t.\ $\ptheta(\vx)$ covers/dominates the data distribution, 
max epoch and any convergence criteria.}
\While{$\vtheta$ not converged within max epoch}{
\emph{Sleep phase: train gradient model}\\
~ Sample $\{\vz_n,\vx_n\}_{n=1}^{N} \sim \ptheta$ \\
~\uIf{$p(\vx|\vz)$ is not in exponential family}{
Find  $\Jtr(\cdot)$ by computing $\valpha_{\vtheta,\vgamma}$ in \eqref{eq:krr_J}\\
}
~\uElse{
Find $\htr^\nat(\cdot)$ and $\htr^\Psi(\cdot)$ similar to \eqref{eq:krr_J}\\
$\Jtr(\cdot)= \htr^\nat(\cdot)\cdot \suff(\vx) - \htr^\Psi(\cdot)$ 
in \eqref{eq:expfam_autodiff_grad}\\
}
~~~~\\
\emph{Sleep phase: update $\vgamma$}\\
~ Sample $\{\vz_l',\vx_l'\}_{l=1}^L \sim \ptheta$ \\
~ Compute $d_l:= \log \ptheta(\vz,\vx) $\\
~ Compute $\gE_\vgamma = \frac{1}{L}\sum_{l=1}^L (\Jtr(\vx_l') - d_l)^2$\\
~ Update $\vgamma\propto\nabla_\vgamma \gE_\vgamma$\\
\emph{Wake phase: update $\vtheta$}\\
~ Sample $\{\vx^*_m\}_{m=1}^M \in \gD$ \\
~ ${\bar{J}_\vtheta}=\frac{1}{M} \sum_i^M \Jtr(\vx^*_m)$\\
~ Update $\vtheta\propto \ntheta {\bar{J}_\vtheta}$ \\
}
\Output{$\vtheta$}
\caption{Amortised learning by wake sleep}
\label{alg}
\end{algorithm}

\subsection{Kernel Structure and Learning}\label{sec:gamma_parameter}
The kernel $k_\vomega$ used in the gradient model affects how well $\Dxt$ is estimated.
It can be made more flexible by augmenting with a neural network as in
\citep{wilson2016deep,WenliangEtAl2019} 
$$
k_\vomega (\vx,\vx') = \varkappa_{\vsigma}(\vpsi_\vv(\vx),\vpsi_\vv(\vx'))
$$
where $\varkappa_\vsigma$ is a standard kernel (e.g.\ exponentiated-quadratic) with parameter $\vsigma$ (e.g.\ bandwidth),
and $\vpsi_\vv$ is a neural network with parameter $\vv$, so $\vomega = \{\vsigma, \vv \}$.
Other details of the kernel structure are described in \cref{sec:kernel_details}.

The gradient model parameter $\vgamma=\{\vomega,\lambda\}$ can be learned to further minimise the MSE in \eqref{eq:autodiff_loss}
using a scheme of cross-validation by gradient descent \citep{WenliangEtAl2019}.
Specifically, we generate two sets of sleep samples from $\ptheta$; 
we use one set to compute $\valpha_{\vtheta, \vgamma}$ in closed form; 
then, on the other set $\{(\vz'_l,\vx'_l)\}_{l=1}^L$, we compute the MSE between the estimator 
$\Jtr(\vx_l')$ 
and the ground truth value $\log\ptheta(\vz_l',\vx_l')$, and minimise this by gradient descent on $\vgamma$.
The full ALWS procedure is presented in \cref{alg}.

\subsection{Dealing with Covariate Shift}\label{sec:covar_shift}
The gradient model is to be used to estimate $\Dxt$ 
on $\vx^*$ drawn from an underlying data distribution $p^*$,  
but it is trained using sleep samples from $\pthetat$.
This mismatch in input data distribution for training and evaluation
is known as covariate shift 
\citep{ShimodairaShimodaira2000Improving}.
Here, to ensure that the gradient model performs reasonably well on
$p^*$, we initialise $\ptheta(\vx)$ to be  
overdispersed relative to $p^*$ by setting a large noise in $\ptheta(\vx|\vz)$.
Since ML estimation minimises $\KL[p^*\|p_\vtheta]$, which 
penalises a distribution $p_\vtheta$ that is narrower than $p^*$, we expect
the noise to continue to cover the data before the model is well trained.
For image data only, we also apply batch normalisation in $\vpsi_\vw$ of the kernel.
We find these simple remedies to be effective, though other more principled  
methods, such as kernel mean matching \citep{GrettonSchoelkopf2009Covariate}
and binary classification \citep{GutmannHyvaerinen2010Noise, GoodfellowEtAl2014},
may further improve the results.

\begin{figure}[t]
    \centering
    \includegraphics[width=0.9\columnwidth]{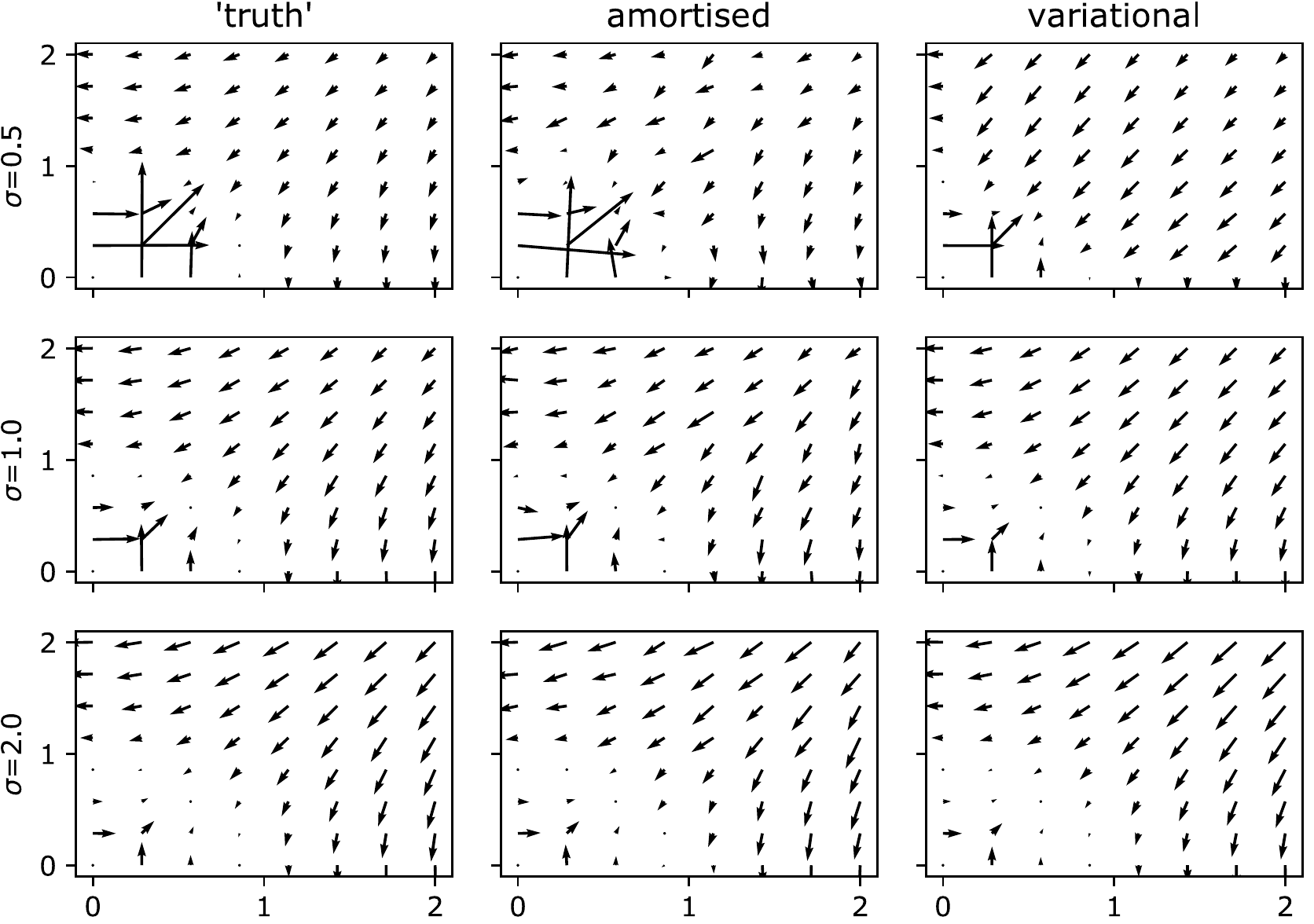}
    \caption{Gradient estimated using amortised learning and variational inference. 
    The true gradients are approximated by importance sampling.\vspace{-0.0cm}}
    \label{fig:gradients}
\end{figure}

\begin{figure}[t!]
    \centering
    \includegraphics[width=\columnwidth]{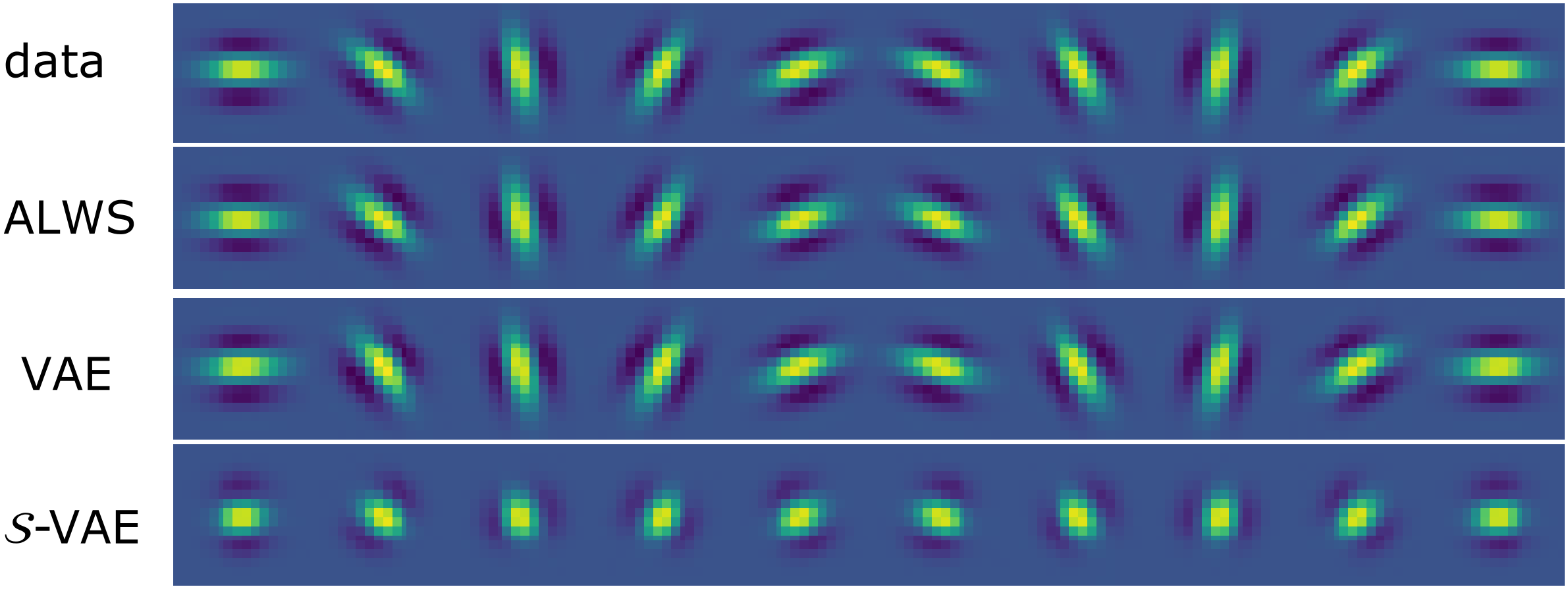}
    \includegraphics[width=\columnwidth]{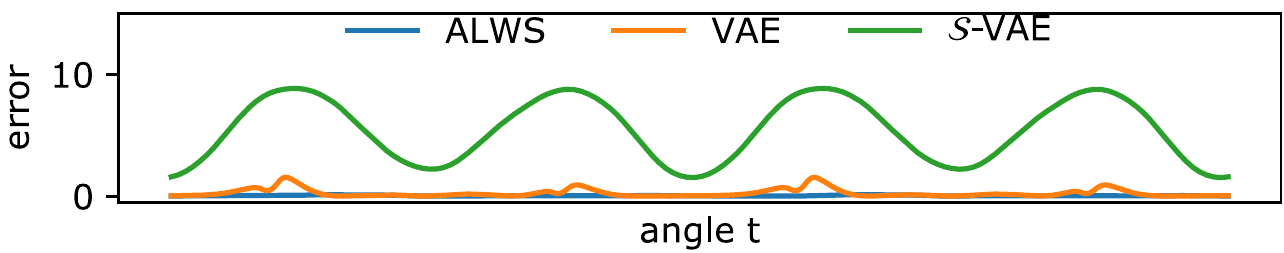}
    \caption{Learning to generate Gabor filters given a 1-D circular uniform prior. 
    Top images show samples generated by latents separated by fixed rotation on the circle. 
    For VAE, a 2-D Gaussian prior was used, and the images are generated by latents on the unit circle.
    $\mathcal{S}$-VAEs cannot reliably learn the filters. 
    The errors below show the squared distance between generated images and data 
    at each orientation. 
    For each method, an angle offset and direction are chosen to minimise the total error.
    }
    \label{fig:gabor}
\end{figure}

\section{Experiments}

We evaluate ALWS on a wide range of generative models. Details for each experiment 
can be found in \cref{sec:exp_details}. \footnote{Code is at \url{github.com/kevin-w-li/al-ws}}

\subsection{Parameter Gradient Estimation}
First, we demonstrate that KRR
can estimate $\Dxt$ well on a simple toy generative model described by
$$
\evz_1,\evz_2\sim \gN(0,1),~x|\vz\sim\gN(\softplus(\vb \cdot \vz)-\|\vb\|_2^2, \sigma_x^2).
$$
The training data are 100 data points from the model given $\vb=[1,1], \sigma_x=0.1$.
we estimate the gradients of the log-likelihood w.r.t.~$\vb$ evaluated at a grid of $\vb$ by ALWS, 
and compare them to estimates using importance sampling
(``truth'') and a factorised Gaussian posterior that minimises the forward KL for each $\vx$.
For ALWS, we used a Gaussian kernel with a bandwidth equal to the median distance between samples
generated for each $\vb$, and set $\lambda=0.01$.
For variational inference, we assumed a factorised Gaussian posterior for each sample of $\vx$, 
and optimise posterior parameters until convergence.
ALWS tends to estimate better, especially for small $\vb$ (\cref{fig:gradients}). For the smallest $\sigma_x$,
the KRR estimates are noisier, whereas variational inference introduces greater bias.

\subsection{Non-Euclidean Priors}\label{sec:gabor}
The prior $p(\vz)$ may capture special topological structures in the data.
For instance, a prior over the hypersphere can be used to describe 
circular features \citep{DavidsonTomczak2018Hyperspherical, xu2018spherical}.
Training models with such a prior is straightforward using ALWS, 
while learning by amortised inference requires special 
reparameterisation for a posterior on the 
hypersphere, such as the von-Mises Fisher (vMF) used   
in the $\mathcal{S}$-VAE \citep{DavidsonTomczak2018Hyperspherical, xu2018spherical}. 
We fit a model with uniform circular latent and neural-network output:
\begin{gather*}
\vz=[\cos(a),\sin(a)],\quad p(a)= \gU(a;(-\pi,\pi)),\\
p(\vx|\vz) = \gN(\vx;\mathrm{NN}_\vw(\vz), \sigma_x^2\mI),
\end{gather*}
(where $\gU$ is a uniform distribution) on a data set of Gabor wavelets 
with uniformly distributed orientations.
As shown in \cref{fig:gabor}, 
ALWS learns to generate images that closely resemble the training data.
A fixed rotation around the latent circle corresponds to almost a fixed rotation 
of the Gabor wavelet in the image. 
The VAE with a 2-D Gaussian latent also generates good filters given latents
on the circle,
but the length of the filter varies with rotation. 
Surprisingly, $\mathcal{S}$-VAE is not
able to learn on this dataset, the vMF posterior is almost flat for any input image.
This hints at potential optimisation issues with the complicated reparameterisation.
This advantage also extends to 
priors over the hyperbolic space, which are used to capture tree-like
hierarchical structures \citep{NaganoKoyama2019wrapped, MathieuTeh2019Continuous}. 

\begin{figure}[t!]
    \centering
    \includegraphics[width=\columnwidth]{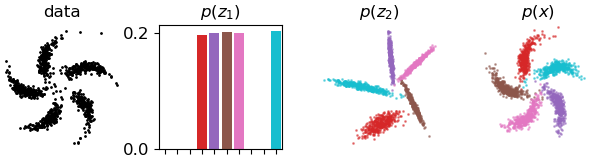}
    \caption{Learning hierarchical model with discrete and continuous latents. 
    From left to right: data sample, component probabilities, samples of 
    the first latent distribution and samples of generated data. Colours correspond
    to different components}
    \label{fig:pinwheel}
\end{figure}

\subsection{Hierarchical Models}\label{sec:pinwheel}
Rich hierarchical structures in the data can be captured with multiple layers of latents.
Provided that samples can be drawn from the hierarchical model and the joint 
log-likelihood evaluated, ALWS extends straightforwardly to hierarchies, 
even with mixed discrete and continuous latents.
The pinwheel distribution \citep{johnson2016composing,LinKhan2018Variational}
has five clusters of distorted Gaussian distributions (\cref{fig:pinwheel}), 
and can be described by the following model:
\begin{gather*}
    p(\vz_1)         = \mathrm{Cat}(\vz_1;\vm),~~
    p(\vz_2|\vz_1=k) = \gN(\vz_2;\vmu_k, \mSigma_k), \\
    p(\vx|\vz_2)     = \gN(\vx;\text{NN}_\vw(\vz_2), \mSigma_x),
\end{gather*}
where $\mathrm{Cat}$ is the categorical distribution.
The parameters are the logits $\vm$ in 10 dimensions, the means and covariance matrices of the component distributions
$\{\vmu_k, \mSigma_k\}_{k=1}^{10}$, 
the weights $\vw$ in $\mathrm{NN}$, and the diagonal covariance $\mSigma_x$.
The logits $\vm$ are penalised according to a Dirichlet prior, 
and $\{\vmu_k, \mSigma_k\}_{k=1}^{10}$ by a normal-Wishart prior.
After training with ALWS, the categorical distribution correctly identifies the five components, 
and the generated samples match the training data. 
We compare these samples with those reconstructed 
from a Bayesian version of the model trained by 
structured inference network (SIN) \citep{LinKhan2018Variational}\footnote{\url{github.com/emtiyaz/vmp-for-svae}}.
A three-way maximum mean discrepancy (MMD) test \citep{BounliphoneGretton2016Test}
finds that samples from the two models are equally close to the training data 
($p=0.514$, $N=1,000$ samples).
Details are in \cref{sec:pinwheel_details}.

\begin{figure}[t]
    \centering
    \includegraphics[width=1.0\columnwidth]{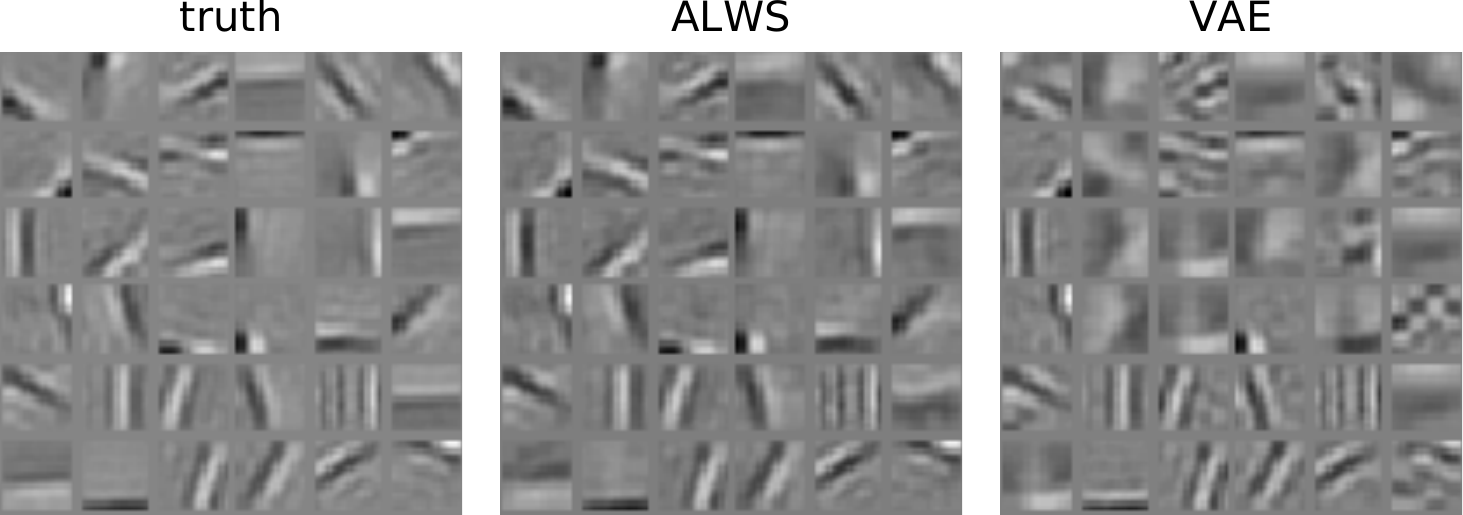}
    \caption{Feature identification. 
    Left, true basis used to generate images.
    Middle, basis recovered by ALWS.
    Right, basis recovered by VAE.
    The filters are arranged according to correlations with the true basis.
    }
    \label{fig:ica}
\end{figure}
\subsection{Feature Identification}

\paragraph{Independent Components.} Learning informative features from complex data can benefit downstream tasks. 
We use ALWS to identify features from data generated by 
\begin{align*}
    p(z_i)=\mathrm{Lap}(z_i;0,1), \quad
    p(\vx|\vz)=\gN(\vx;\mW\vz,\sigma^2I),
\end{align*}
where $\mathrm{Lap}$ is the Laplace distribution, $\sigma=0.1$ and 
basis $\mW$ contains independent components of natural images \citep{HaterenSchaaf1998}
found by the FastICA algorithm \citep{HyvaerinenOja2000Independent}.
Since this model is identifiable, we perform model recovery from a random initialisation of $\mW$ 
using ALWS and compare with a VAE.
ALWS clearly finds better features, as shown in \cref{fig:ica}. 
On generated samples, a three-way 
MMD test favours ALWS over the Laplace-VAE ($p<10^{-5}$) based on $10,000$ samples. 
Details are in \cref{sec:ica_details}.

\begin{figure}
    \centering
    \includegraphics[width=\columnwidth]{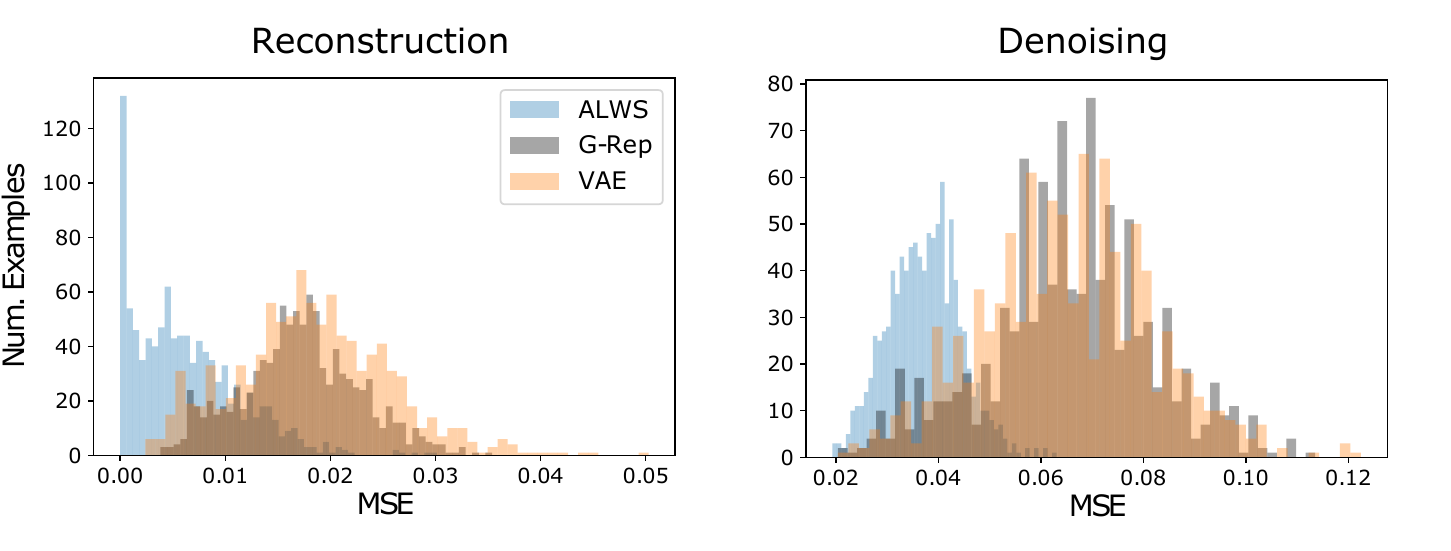}
    \includegraphics[width=\columnwidth]{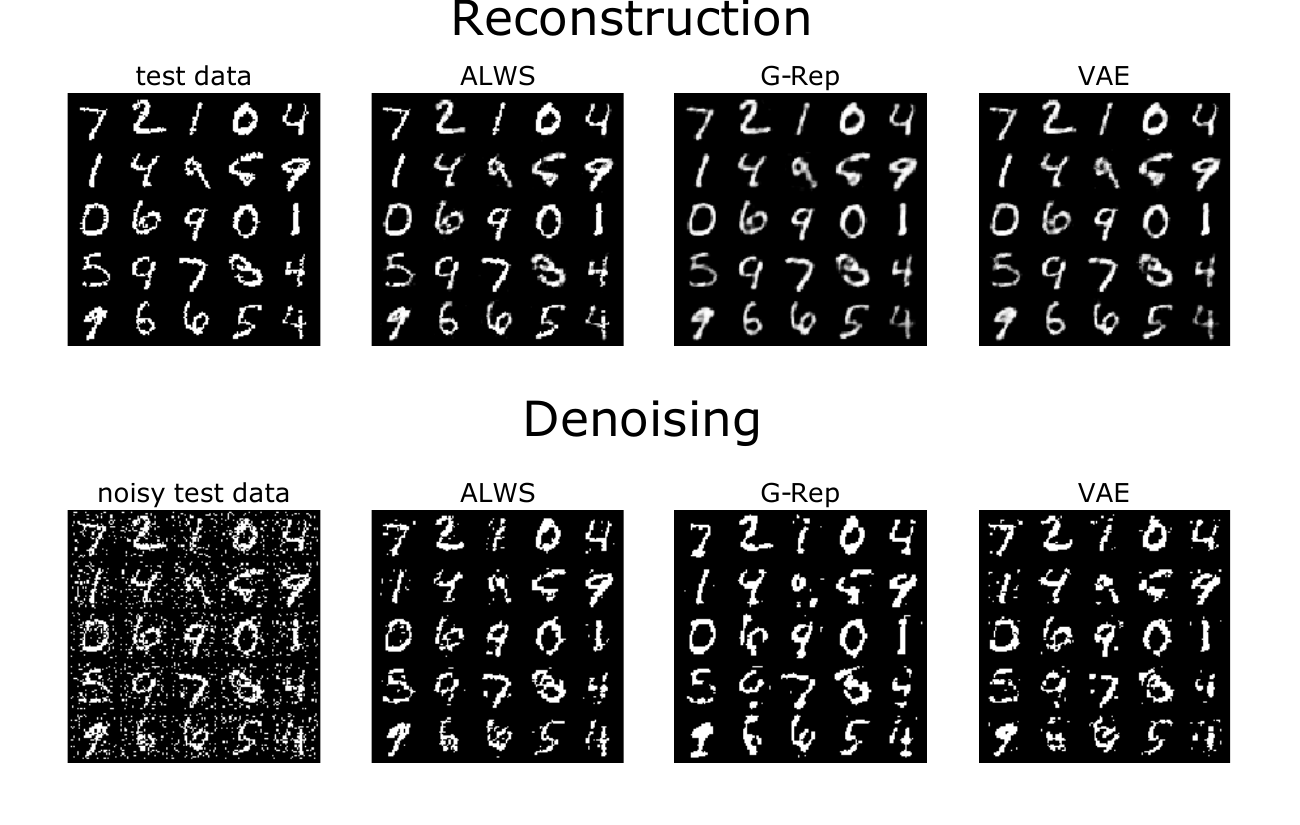}
    \caption{Beta-Gamma Matrix Factorisation. 
    Top, mean squared error 
    across 1,000 test inputs compared to G-Rep and VAE.
    Bottom, examples of real data, reconstructed and denoised samples.
    }
    \label{fig:mf}
\end{figure}

\paragraph{Matrix Factorisation.}\label{sec:bgmf_details}
A more accurate data model may improve performance on a downstream 
task that relies on inference of associated latent variables.
Following \citep{RuizBlei2016Generalized}, we test post-learning inference on a 
probabilistic non-negative matrix factorisation model:
\begin{gather*}
p(z_i)=\gU(z_i;0,1), \quad
p(x_i|\vz)=\operatorname{Bernoulli}\left(x_i;\bar{x}_i\right)\\
\bar{x}_i = \operatorname{sigmoid}\left(\vw_i\cdot\textrm{logit}(\vz) + b_i\right).
\end{gather*}
For each element of each $\vw_i$, we place a penalty consistent 
with a $\textrm{Gamma}(w;0.9,0.3)$ prior on each entry and learn $\mW$ and $\vb$. 
We include $\vb$ to the model trained by ALWS as it prevents samples with opposite 
colour polarity to be generated, which creates a more severe covariate shift
that harms the gradient model.
We evaluate the models on reconstructing and denoising handwritten 
digits from the binarised MNIST dataset. 
To recover the original image given a clean or noisy $\vx^*$, 
we generate $\vx$ given the posterior mode found by maximising 
$\log p(\vz,\vx^*)$ over $\vz$.
We compare with a Bayesian version of the model 
trained by generalised reparameterisation \citet{RuizBlei2016Generalized}
and a VAE-like model in which the decoder has the generative 
structure as above and the posterior is a reparametrised Beta 
distribution.
The results for both tasks are depicted 
in \cref{fig:mf}. 
The leftmost panels show the histograms of MSE on 1\,000 test images, 
and the other panels show examples of 25 test images and reconstructions by each method. 
ALWS achieved significantly lower error  
($p<10^{-10}$ for both a two-tailed $t$-test and a Wilcoxon signed-rank test).

\subsection{Neural Processes}
The neural process (NP) \citep{garnelo2018neural} is a model that learns to infer over functions. Conceptually, 
the computational goal of NPs is similar to predictive inference in 
Gaussian Processes,  
but without defining an explicit prior over functions.
We review NPs in more detail and illustrate how they can be trained by ML using ALWS 
in \cref{sec:np_details}. 
We compared ALWS with the original variational learning method on a toy problem.
NP trained by ALWS produces better prediction and uncertainty estimates on test inputs. 
See \cref{fig:np} in \cref{sec:np_details}.

\subsection{Dynamical Models}\label{sec:dynamical}
In fields such as biology and environmental science, 
the behaviour of complex systems is often
described by simulation-based dynamical models.
Estimating parameters for these models from data
is crucial for prediction and policy-making.
\citep{LintusaariCorander2016Fundamentals, sunnaaker2013approximate, kypraios2017tutorial}
\begin{figure}[t]
    \centering
    \includegraphics[width=0.95\columnwidth]{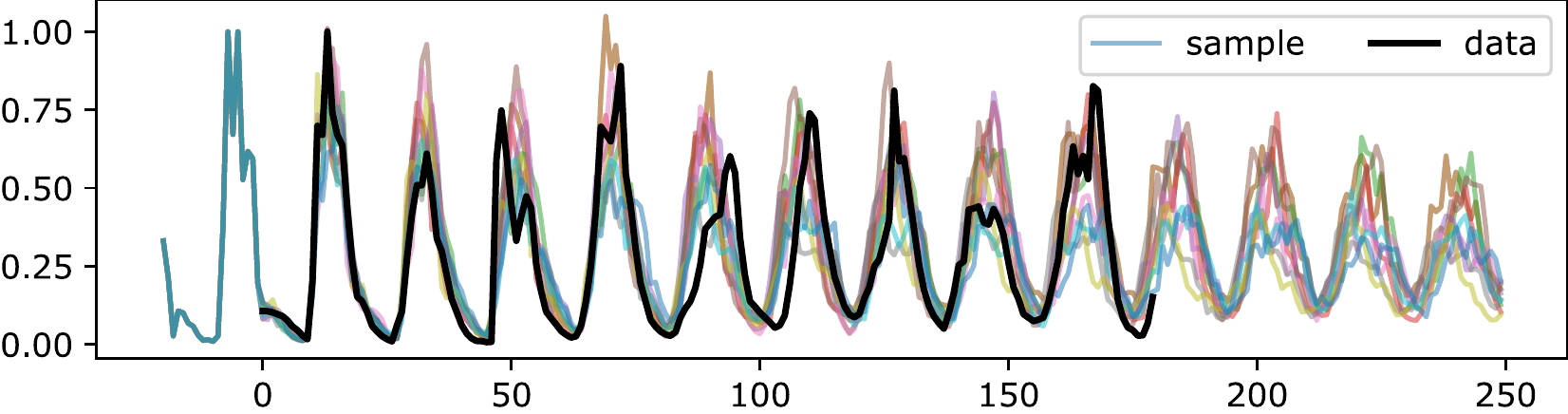}
    \caption{Modelling blowfly population time series. 
    Black, training data. Coloured, samples for an extended time period
    drawn from the trained model.}
    \label{fig:eco}
\end{figure}

A dynamical model can be expressed, in discrete time, as
\begin{gather*}
    \vz_t=\vl_\vtheta(\vz_{1:t-1},\vx_{1:t-1},\vu_t,\vepsilon_t),\quad\vx_t=\vo_\vtheta(\vz_t) + \ve_t
\end{gather*}
where $\vl_\vtheta$ describes a latent process that can depend on 
a control input $\vu_t$,
a noise source $\vepsilon_t$ 
and the history of latents $\vz_{1:t-1}$ and measurements $\vx_{1:t-1}$.
The function $\vo_\vtheta$ maps the latent $\vz_t$ to 
measurement with noise $\ve_t$.
For ALWS, we need that 
$\ptheta(\vz_t,\vepsilon_t|\vz_{1:t-1},\vx_{1:t-1},\vu_t)$ and 
$\ptheta(\vx_t,\ve_t|\vz_t)$ 
are tractable so that $\ntheta\log p(\vz_{1:T},\vx_{1:T})$ can be evaluated, where $T$ is 
the length of the data.
However, learning using approximate inference may be challenging due to 
complex dependencies between latent variables and across time.

Here, we fit the parameters of two dynamical models:  
the Hodgkin-Huxley (HH) model \citep{PospischilDestexhe2008Minimal} 
on the membrane potential of a simulated neuron,
and an ecological model (ECO) on blowfly data \citep{WoodWood2010Statistical}.
The HH equations describe the membrane potential and 
three ion-channel state variables of a neuron that follow
complicated nonlinear transitions. 
Details of the experiment are in \cref{sec:dynamic_details}. Results in \cref{fig:hh}
show that the trained model can not only reproduce the training data well 
but also predict the response given new inputs $\vu_t$.
ECO describes nonlinear and non-Gaussian 
dynamics and has discrete and continuous latent variables. 
Fitting ECO on blowfly data was used to validate approximate Bayesian computation (ABC) methods 
\citep{ParkSejdinovic2016K2-ABC}. 
The model trained with ALWS can simulated sequences very close to data \cref{fig:eco}, 
and are visibly closer than sequences from the model trained with ABC
\citep[Figure 2b]{ParkSejdinovic2016K2-ABC}. 
\begin{figure*}[ht!]
    \centering
    \includegraphics[width=\textwidth]{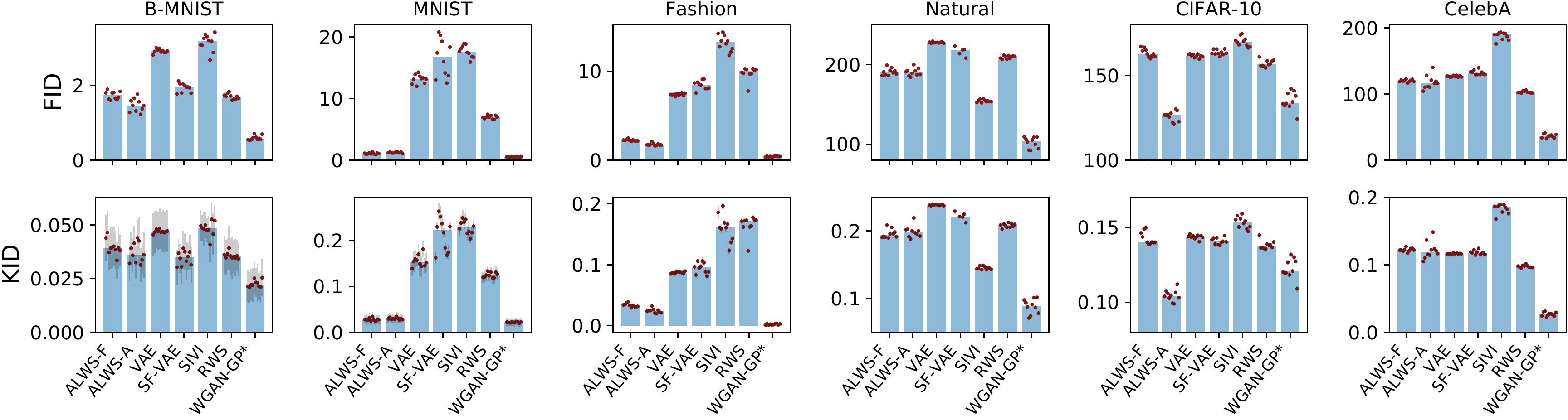}
    \caption{
    FID and KID scores (lower is better) for different datasets and methods. 
    Red dot is the score for a single run.
    Bars are medians of the dots for each method.
    Short bars on KID dots shows standard error of the estimate.
    All models are trained for 50 epochs.
    }
    \label{fig:real_kid}
\end{figure*}

\subsection{Sample Quality}\label{sec:benchmark}
Finally, we train deep models of images and test sample quality. 
We chose six benchmark datasets: the binarised and original 
MNIST \cite{LeCunEtAl1998} (B-MNIST and MNIST, respectively), 
fashion MNIST (Fashion) \citep{XiaoEtAl2017}, 
natural images (Natural) \citep{HaterenSchaaf1998}, CIFAR-10 \citep{Krizhevskyothers2009Learning}
and CelebA \citep{LiuTang2015Deep}.
The original un-binarised MNIST is known to be difficult 
for most VAE-based methods \citep{LoaizaGanemCunningham2019}.
Natural images consist of grey-scale images from natural scenes.
All images have size $32 \times 32 $ with colour channels.
For ALWS, we test two variants.
In ALWS-F, gradient model parameters $\vgamma$ are fixed. 
In ALWS-A, $\vgamma$ is adapted as described in 
\cref{sec:gamma_parameter} except for $\lambda$ which is fixed at $0.1$. 
Fixing $\lambda$ improved quality for the higher-dimensional CIFAR-10 and CelebA, 
but lowered quality for Natural and did not affect much on the other datasets.

We compare these methods with four other approaches: 
the vanilla VAE \citep{KingmaWelling2014}, 
VAE with a Sylvester (orthogonal) flow as an inference network 
\citep{BergWelling2018Sylvester} (Syl-VAE)%
\footnote{\httpsurl{github.com/riannevdberg/sylvester-flows}}, 
semi-implicit variational inference \citep{YinZhou2018Semi} (SIVI)\footnote{\httpsurl{github.com/mingzhang-yin/SIVI}},
and reweighted wake-sleep \citep{BornscheinBengio2015Reweighted}. 
Each algorithm has the same generative network architecture as in DCGAN\footnote{\httpsurl{pytorch.org/tutorials/beginner/dcgan_faces_tutorial.html}}
with the last convolutional layer removed.
We also run WGAN-GP \citep{GulrajaniCourville2017Improved}%
\footnote{\httpsurl{github.com/caogang/wgan-gp}} for reference, 
although it is not trained by ML methods.
Each algorithm is run for 50 epochs ten times with different initialisations, 
except for SIVI where we trained for 1000 epochs with a lower learning rate for stability.
To test the generative quality, we compute both the 
Fr\'{e}chet Inception Distance (FID) \citep{HeuselHochreiter2017GANs} and 
Kernel Inception Distance (KID) \citep{BinkowskiGretton2018Demystifying}
on 10,000 generated images.
The results are shown in \cref{fig:real_kid}.
According to FID, ALWS-A is the best ML method for binarised MNIST, Fashion, and CIFAR-10. 
Notably, both ALWS-A and ALWS-F have much smaller FID and KID on MNIST and Fashion than other ML methods.
WGAN-GP did not produce a good score on CIFAR-10 within 50 epochs but becomes 
the best model for all datasets with further training.
Samples are shown from \cref{fig:real_mnist_0_conv} to \cref{fig:real_celeb_0_conv} in \cref{sec:real_details} with additional 
experiments to show the effectiveness of ALWS.

\section{Related Work}
\subsection{Amortised Variational Inference}

Using $\gF(q,\vtheta)$ as the objective for learning $\vtheta$,
the gradient for $\vtheta$ is given by an intractable posterior expectation. 
The large majority of learning algorithms based on amortised variational inference use Monte Carlo estimators for the gradient. 
The Variational auto-encoder (VAE) \citep{KingmaWelling2014,RezendeEtAl2014} 
parametrises $\qphi(\vz|\vx)$ by simple distributions
using reparameterised samples to obtain gradients for $\vpsi$.
Approximate posteriors may also be incorporated into tighter bounds on $\lpmargin$ by reweighting \citep{BurdaEtAl2016, BornscheinBengio2015Reweighted,LeEtAl2019}, although with some loss of gradient signal \citep{RainforthEtAl2018}.
More expressive forms of $\qphi$ can be formed by invertible transformations (normalising flows) 
\citep{RezendeMohamed2015,KingmaEtAl2016,BergWelling2018Sylvester}) that allow $\sH[\qphi]$ to be
computed easily, 
or by non-invertible mappings (implicit variational inference),
which requires estimating $\sH[\qphi]$ or its gradient w.r.t. $\vphi$
\citep{ShiEtAl2018,LiTurner2018, YinZhou2018Semi,huszar2017variational}.
Reparametrising posterior samples may require nontrivial methods
\citep{JangPoole2017Categorical,VahdatAndriyash2018DVAE++,RolfeRolfe2017Discrete, RuizBlei2016Generalized,FigurnovMnih2018Implicit}.
On the other hand, amortised learning focuses exclusively on estimating 
the gradient for ML learning, making no assumptions on the type of latent variables.

Our approach is related to at least two other algorithms inspired by the original Helmholtz machine (HM)
\citep{dayan1995helmholtz, hinton1995wake}.
The distributed distributional code HM (DDC-HM) \citep{vertes2018flexible} 
represents posteriors by expectations of pre-defined and finite nonlinear features,
which are used to approximate $\Dxt$ by the linearity of expectation.
ALWS differs from DDC-HM in two ways.
First, our gradient model integrates the inferential model
and the linear readout for $\Dxt$ in DDC-HM
 using adaptive and more flexible KRR.
Second, using \eqref{eq:exact_grad_autodiff} avoids explicit computation of 
$\nlptheta$ and makes ALWS easily applicable to more complex generative models.
Reweighted wake-sleep (RWS) \citep{BornscheinBengio2015Reweighted} 
addressed covariance shift by training an inferential model to increase the likelihood of not only sleep $\vz$
given sleep $\vx$ as in the HM, but also weighted posterior samples given data $\vx^*$.
ALWS does not make assumptions about the posterior distributions, and we 
found that simple strategies mitigated covariate shift in practice, but 
this is a point that deserves further investigation.

\subsection{Training Implicit Generative Models}

Implicit generative models, 
including generative adversarial networks (GANs) \citep{GoodfellowEtAl2014}
and simulation-based models considered by approximate Bayesian computation (ABC)
\citep{tavare1997inferring,marin2012approximate},
do not have an explicitly defined likelihood function
but can be trained using simulated data.
Amortised learning requires an explicit joint likelihood function $\ptheta(\vx,\vz)$, 
but can also train simulation-based generative models (\cref{sec:dynamical}).
In GANs, the generator is improved by a 
discriminator that is concurrently trained to tell apart real and generated samples.
The approach is able to synthesise high-quality samples in high dimensions.
However, the competitive setting can be problematic for convergence, and 
the discriminator needs to be carefully regularised to be less effective at 
its own task but more informative to the generator.
\citep{ArjovskyEtAl2017,GulrajaniCourville2017Improved,ArbelGretton2018gradient,MeschederNowozin2018Which}.
In amortised learning, a better gradient model always 
helps when training the generative model. 
Importantly, amortised learning can directly train real-world simulators 
for which samples of $\vx$ are not differentiable w.r.t.\ $\vtheta$, 
such as the Galton board, where GANs are not directly
applicable.

Rather than performing maximum likelihood estimation,
ABC estimates a posterior of $\vtheta$ 
using simulated data and a chosen prior on $\vtheta$. 
Amortised learning can be seen as maximum likelihood learning 
based on simulations,
since the gradient model is trained using data from the generative model.
In particular, ALWS is similar to Kernel-ABC \citep{nakagome2013kernel} 
in which the posterior is found by weighting prior samples using KRR
on pre-defined summary statistics.
The kernel recursive ABC \cite{kajihara2018kernel} iteratively 
updates the prior over $\vtheta$ by herding from 
a kernel embedding \citep{SongFukumizu2009Hilbert} of the posterior,
converging to a maximum likelihood solution.
ALWS does not maintain a distribution of $\vtheta$, 
but iteratively updates them by gradient methods so that the model distribution approaches 
the data distribution.
Also, ALWS performs well even when the number of parameters is large
for which traditional ABC methods are likely to be expensive. 

\section{Discussion} 

Direct estimation of the expected log-likelihood and its gradient in a latent variable model
circumvents the challenges and issues posed by explicit approximation of posteriors. 
The KRR gradient model is consistent, easy to implement, and avoids the need for explicit 
computation of derivatives.
However, we observe the following issues with the current instance 
of amortised learning. 
First, its computational complexity limits the number of 
sleep samples that can be used to train the gradient model
and thus the quality of the approximation.
Techniques such as random feature- and Nystrom-approximations could 
make KRR more efficient.
Second, 
the KRR prediction is a linear combination of the set
$\{\nlptheta(\vz_n,\vx_n)\}_{n=1}^N$,
but the true gradient function, which can be much higher-dimensional than $N$,
may lie outside this span---an issue that might be compounded by
covariate shift.
Further, hyper-parameter learning using the meta-learning method 
described in \cref{sec:gamma_parameter} improves the estimation of $\Jtheta$
rather than $\nJtheta$, which might explain why adapting $\lambda$ 
on some tasks worsens the results.
Therefore, alternative amortised learning models may be worth future exploration.
Nonetheless, we have found here that ALWS based on KRR provides accurate parameter estimates in
many settings where approximate inference-based approaches appear to struggle.

ALWS can be extended to training generative models of 
other types of data, such as graphs, 
as long as an appropriate kernel is used.
Another useful extension is to train conditional generative models,
which we explored briefly in the neural processes experiment. 
In this case, 
the gradient model needs to depend on any 
conditioning variables (or sets).
Finally, while we used LSR to approximate the gradient of 
the model w.r.t\ $\theta$,
other useful quantities could also be estimated in a 
similar fashion \citep{BrehmerCranmer2020Mining}.

\section*{Acknowledgements}
We thank Arthur Gretton, Sebastian Nowozin, Jiaxin Shi and Eszter V\'ertes 
for helpful discussions; we thank Ferenc Husz\'ar for discussion and comments on an earlier draft.

\bibliography{ref}
\bibliographystyle{icml2020}

\appendix
\onecolumn

\section{Mathematical details}

\subsection{Solving mean squared error for conditional expectations}\label{sec:mse_for_mean_proof}

Given $\vx,\vy\sim\rho(\vx,\vy)$, 
we want to find an estimator in some space $\gF$ of the posterior mean function $\vf_{\rho}: \vx\mapsto \E{\rho(\vy|\vx)}{\vy}$.
Assuming that $\gF$ is contained in $\gL_\rho^2$, the class of squared-integral functions under $\rho(\vx)$, 
and that $\vy$ has finite $l$-2 norm under $\rho(\vy)$, 
a natural cost function to learn $\vf$ is the expected squared $l$-2 distance 
$$
L_{E}(\vf):=\E{\rho(\vy, \vx)}{\left\|\vf(\vx)-\vy\right\|_2^2}
= \E{\rho(\vx)}{\E{\rho(\vy|\vx)}{\left\|\vf(\vx)-\vy\right\|_2^2}}.
$$
By Jensen's inequality,
$$
L_{E} (\vf) \le \E{\rho(\vx)}{\left\|\vf(\vx)-\E{\rho(\vy|\vx)}{\vy}\right\|_2^2} =L_{R}(\vf).
$$
This shows that the MSE is an upper bound on the expected $l$-2 distance between $\vf(\vx)$ and 
the posterior mean $\E{\rho(\vy|\vx)}{\vy}$. 
Further, the minimum of $L_{R}$ is attained at an $\vf$ that also minimises $L_{E}$. 
This can be shown through a simple decomposition
\begin{align*}
L_{E}(\vf) &= \E{\rho(\vx)}{\E{\rho(\vy|\vx)}{\left\|\vf(\vx)-\vy\right\|_2^2}}\\
&= \E{\rho(\vx)}{\|\vf(\vx)\|_2^2-\vf(\vx) \cdot \E{\rho(\vy|\vx)}{\vy}+\E{\rho(\vy|\vx)}{\|\vy\|_2^2}}\\
&\stackrel{(1)}{=} \E{\rho(\vx)}{\|\vf(\vx)\|_2^2-\vf(\vx) \cdot \E{\rho(\vy|\vx)}{\vy}+\left\|\E{\rho(\vy|\vx)}{\vy}\right\|_2^2 + 
                                                                              \mathrm{Tr}\left[\mathbb{C}_{\rho(\vy|\vx)}[\vy]\right] }\\
&= \E{\rho(\vx)}{\E{\rho(\vy|\vx)}{\left\|\vf(\vx)-\E{\rho(\vy|\vx)}{\vy}\right\|_2^2}} + 
                 \E{\rho(\vx)}{\mathrm{Tr}\left[\mathbb{C}_{\rho(\vy|\vx)}(\vy\right] }\\
&= L_{R}(\vf) + \textrm{term independent of $\vf$}
\end{align*}
where $\mathbb{C}_{p}$ is the covariance under $p$. Equality $(1)$ holds because 
$$
\E{p}{\|\va\|_2^2} 
= \E{p}{\sum_i a_i^2} 
= \sum_i\E{p}{a_i^2} = \E{p}{a_i}^2+\sum_i \mathbb{V}_{p}[a_i] 
= \left\|\E{p}{\va}\right\|_2^2 + \mathrm{Tr}\left[\mathbb{C}_{p}[\va]\right]
$$
for any $\va\in$ in $\gL^2_p$.
So $L_R(\vf)$ is equal to $L_{E}(\vf)$  up to a constant that depends only on $\rho$ but 
not $\vf$. 

\subsection{Boundedness of the gradient function}\label{sec:l2p}

To learn $\vy(\vx)=\E{\pthetat(\vz|\vx)}{\nlpjoint}\evalt$ using regression as above, 
the target needs to be square-integrable under $\pthetat(\vx)$, i.e. $\vy(\vx)\in \gL_p^2$.
Common likelihood functions are in the exponential family and has
$\nlpjoint=\ntheta\nat(\vz) \suff(\vx) - \ntheta\lnorm(\vz)$. 
Thus, it suffices to check the $\gL_p^2$ integrability of the gradient in terms of these functions. 
We sketch below that this is indeed the case for common choices of model architectures. 

As a simple example, consider a model 
\begin{equation}\label{eq:gaussian_model}
\ptheta(\vz)=\gN(\mathbf{0}, \mI), \qquad \ptheta(\vx|\vz)=\gN(\mathrm{NN}_\vw(\vz), \mSigma). 
\end{equation}
where $\mI$ is the identity covariance matrix, 
$\textrm{NN}_\vw$ is a neural network with weights $\vw$ 
and $\mSigma$ is a diagonal matrix. 
Note that in this case, one has that 
$$
\lnormt(\vz)=-\frac{1}{2}\|\vz\|_2^2 - \frac{1}{2} \log |\mSigma| + \text{constant}, 
\quad 
\natt(\vz)=[\mSigma^{-1}\textrm{NN}_\vw(\vz), -\frac{1}{2}\mSigma^{-1}],
\quad 
\vtheta=\{\vw,\mSigma\},
\quad 
\suff(\vx)=[\vx,\vx\vx^T].
$$

Further, assume that 
\begin{enumerate}
    \item The neural network $\mathrm{NN}_{\vw}(\vz)$ is Lipschitz and $\vw$-differentiable almost everywhere, such as 
    one that is composed of linear projections followed by Lipschitz nonlinearities (e.g., ReLU).
    \item $\E{\ptheta(\vz)}{\|\mathrm{NN}_{\vw}(\vz)\|_2^2} < \infty$.
    \item Spectral norm of weights $\mW$ in each layer of $\mathrm{NN}_{\vw}(\vz)$ is bounded above by a positive constant.
    \item The diagonal elements of $\mSigma$ are bounded below by some constant. 
\end{enumerate}
The first and second assumptions are mild and satisfied by NNs with ReLU activations. 
The third and fourth conditions limit the ranges of the parameter values, which can be imposed by clipping or through appropriate parametrisation. 

The second and fourth conditions make the gradients of $\lnormt(\vz)$ and $\natt(\vz)$ w.r.t. $\mSigma$ bounded; thus, we will demonstrate the integrability of the gradients w.r.t. the neural network parameter $\vw$. 

\textbf{First term $\E{\ppost}{\nabla_{\vtheta}\natt(\vz)}\suff(\vx)$}

Multiple applications of the Cauchy-Schwartz inequality yields 
\[
\E{\ptheta(\vx)}{\bigl\|\E{\ppost}{\nabla_{\vw}\natt(\vz)}  \suff(\vx)\bigr\|^2} \leq 
\sqrt{\E{\ptheta(\vx)}{\|\E{\ppost}{\nabla_{\vw}\natt(\vz)}\|_2^4\|}}\sqrt{\E{\ptheta(\vx)}{\|\suff(\vx)\|_2^4}}.
\]
By our assumption, $\mathrm{NN}(\vz)$ is Lipschitz w.r.t.\ $\vw$ and the gradient $\nabla_{\vw}\natt(\vz)$ is bounded as, for $C_0, C_1>0$, $\left\|\nabla_{\vw}\natt(\vz)\right\|_2 \le C_0 +C_1 \|\vz\|_2$.
This can be proved by writing out $\nabla_{\vw}\textrm{NN}_\vtheta(\vz)$ using the chain rule, which will be a series of product
involving $\mW$ in each layer and derivative of Lipschitz functions, and applying the first two conditions above.
Thus, we have 
\begin{align*}
\E{\ptheta(\vx)}{\left\|\E{\ppost}{\nabla_{\vw}\natt(\vz)}\right\|_2^4 } 
&\le \E{\ptheta(\vx)}{\E{\ppost}{\left\| \nabla_{\vw}\natt(\vz)\right \|_2^4}}\\
& \le \E{\ptheta(\vz)}{(C_0+C_1\left\|\vz \right\|_2)^4} < \infty \\
\end{align*}
as the prior $\ptheta(\vz)$ is a standard Gaussian. 

The integrability of $\suff(\vx)$ is equivalent to the finiteness of the corresponding moments of $\ptheta(\vx)$. 
By Lemma \ref{lem:gaussian_exp_tail}, the marginal $\ptheta(\vx)$ has exponential tails, and thus the moments are finite. 

\textbf{Second term $\ntheta\lnorm(\vz)$}

$$
\E{\pthetat(\vz)}{\left\| \E{\ppostt}{\ntheta\lnormt(\vz)}\right\|_2^2} 
\le\E{\pthetat(\vz)}{\E{\ppostt}{\left\| \ntheta\lnormt(\vz)\right\|_2^2}} 
= \|\mSigma^{-1}\|_2^2 < \infty
$$ 
where we have applied Jensen's inequality. Therefore, $\E{\ptheta(\vz|\vx)}{\ntheta\lnorm(\vz)}$ is a finite constant and thus 
in $\gL_p^2$.

Therefore, for the generative model defined in \eqref{eq:gaussian_model}, 
the desired target $\vy(\vx)=\E{\pthetat(\vz|\vx)}{\nlpjoint}\evalt$ for regression is in $\gL_p^2$, which can be 
approximated arbitrarily well by KRR (see \cref{sec:KRR})) with more sleep samples.
A similar analysis can show that for Bernoulli likelihoods whose logits are parametrised by a Lipschitz neural network, 
the target for the regression is also in $\gL_p^2$,
with logits bounded from above and below.

\subsection{Gradient of the log marginal likelihood w.r.t. parameters} \label{sec:Dx_proof}
To show the result used in \eqref{eq:free_energy_exact_grad}, we start  
from the free energy (ELBO) lower bound on the log-likelihood $\log \ptheta(\vx)$. 
\begin{align*}
\log\ptheta(\vx)  &= \log \frac{\ptheta(\vz, \vx)}{\ptheta(\vz|\vx)}   
                   = \int q(\vz) \log   \left[\frac{q(\vz)}{q(\vz)}
                                            \frac{\ptheta(\vz, \vx)}{\ptheta(\vz|\vx)}
                                        \right]\ud\vz  
                   = \int q(\vz) \log   \left[\frac{\ptheta(\vz, \vx)}{q(\vz)}
                                            \frac{q(\vz)}{\ptheta(\vz|\vx)}
                                        \right]\ud\vz  \\
                  &= \int q(\vz) \log \ptheta(\vz,\vx) \ud\vz 
                        - \int q(\vz) \log q(\vz) \ud\vz 
                        + \KL[q(\vz)\|\ptheta(\vz|\vx)]\\
                  &= \gF(q,\vtheta) + \KL[q(\vz)\|\ptheta(\vz|\vx)], \numberthis \label{free_energy_KL}
\end{align*}
where we have defined 
\begin{equation*}
    \gF(q,\vtheta)= \int q(\vz) \log \ptheta(\vz,\vx) \ud\vz - \int q(\vz) \log q(\vz) \ud\vz 
    =\E{q(\vz)}{\lptheta(\vz,\vx)} + \sH[q].
\end{equation*}
The KL term in \eqref{free_energy_KL} is non-negative and is zero if $q(\vz)=\ptheta(\vz|\vx)$, 
suggesting that 
$$ \log \ptheta(\vx) = \gF(\ptheta(\vz|\vx),\vtheta)$$
Replacing $q(\vz)=\ptheta(\vz|\vx)$ in \eqref{free_energy_KL} and 
take derivative w.r.t.\ $\theta$ gives (assuming all derivatives and expectations exist)
\begin{align*}
    \Dx &:=\ntheta \log \ptheta(\vx) \\
    &=   \ntheta \int \ptheta(\vz|\vx)\log\ptheta(\vz,\vx)\ud\vz
        - \ntheta \int \ptheta(\vz|\vx)\log\ptheta(\vz|\vx)\ud\vz\\
    &=  \int \ntheta \ptheta(\vz|\vx)\log\ptheta(\vz,\vx)\ud\vz 
        + \int \ptheta(\vz|\vx)\ntheta \log\ptheta(\vz,\vx)\ud\vz\\
    &\quad -\int \ntheta \ptheta(\vz|\vx)\log\ptheta(\vz|\vx)\ud\vz
             -\int \ptheta(\vz|\vx)\ntheta \log\ptheta(\vz|\vx)\ud\vz.
             \numberthis \label{eq:four_terms}
\end{align*}
The last term in \eqref{eq:four_terms} is zero since it is the expectation of the score function
$$
\int \ptheta(\vz|\vx) \nabla \log \ptheta (\vz|\vx)\ud\vz 
= \int \ptheta(\vz|\vx) \frac{1}{\ptheta(\vz|\vx)}\ntheta \ptheta(\vz|\vx)\ud\vz 
= \ntheta\int\ptheta(\vz|\vx)\ud\vz=0.
$$
The first and third terms in \eqref{eq:four_terms} combines to give 
$$
\int \ntheta \ptheta(\vz|\vx) \log\frac{\ptheta(\vz,\vx)}{\ptheta(\vz|\vx)}\ud\vz
=\int \ntheta \ptheta(\vz|\vx) \log \ptheta(\vx)\ud\vz 
= \log \ptheta(\vx)\ntheta\int \ptheta(\vz|\vx)\ud\vz
=0.
$$
We are left with only the second term in \eqref{eq:four_terms}
\begin{equation}\label{eq:one_term}
\Dx = \int{\ptheta(\vz|\vx)}{\ntheta\log \ptheta(\vz,\vx)}
=\E{\ptheta(\vz|\vx)}{\ntheta\log \ptheta(\vz,\vx)}
=\ntheta\gF(\ptheta(\vz|\vx),\vtheta).
\end{equation}
To compute the update at the $t$'th iteration with $\vtheta=\vtheta_t$, 
and the expectation above is taken over a fixed posterior distribution $\pthetat(\vz|\vx)$. 
We evaluate the above equation at $\vtheta_t$, giving \eqref{eq:free_energy_exact_grad},
$$
\Dxt:=\Dx\evalt
= \ntheta \E{\pthetat(\vz|\vx)}{\log \ptheta(\vz,\vx)}\evalt
= \ntheta \gF(\ptheta(\vz|\vx),\vtheta)\big\rvert_{\vtheta_t}.
$$
One can also pass $\ntheta$ and its evaluation inside the 
expectation (assuming derivatives exist) to obtain \eqref{eq:exact_grad_manual} 
$$
\Dxt
= \ntheta \E{\pthetat(\vz|\vx)}{\log \ptheta(\vz,\vx)}\evalt
= \E{\pthetat(\vz|\vx)}{\ntheta \log \ptheta(\vz,\vx)\evalt}
$$
which is used for direct gradient estimation.

In fact, once we know the result above, going from the right-hand side to the left is much simpler:
\begin{align*}
\E{\pthetat(\vz|\vx)}{\ntheta \log \ptheta(\vz,\vx)\evalt}
&=\E{\pthetat(\vz|\vx)}{\ntheta \log \ptheta(\vz|\vx)\evalt + \ntheta \log \ptheta(\vx)\evalt}\\
&=\ntheta \E{\pthetat(\vz|\vx)}{\log \ptheta(\vz|\vx)}\evalt  + { \ntheta \log \ptheta(\vx)\evalt}\\
&= 0 + \Dxt.
\end{align*}
Additionally, a quicker and more direct way to obtain \eqref{eq:one_term} uses the ``score trick'' as follows
$$
\nabla\lptheta(\vx) 
= \frac{1}{\ptheta(\vx)}\ntheta\int\pjoint \ud\vz 
= \frac{1}{\ptheta(\vx)} \int \ptheta(\vz,\vx) \ntheta\lpjoint \ud\vz
= \E{\ptheta(\vz|\vx)}{\ntheta\lpjoint}.
$$

\subsection{Miscellaneous results}
\begin{theorem}[Gaussian concentration inequality {\citep[Theorem 5.6]{boucheronConcentrationInequalitiesNonasymptotic2013}}]\label{thm:gaussian_concentration}
Let $X=(X_1,\dots,X_n)$ be a vector of $n$ independent standard normal random variables. Let $f:\mathbb{R}^n\to\mathbb{R}$ denote an $L$-Lipschitz function. Then, all $t>0$, 
\[P\left[f(X)-\mathbb{E}f(X)\geq t\right]\leq e^{-t^2/(2L^2)}.\]
\end{theorem}
\begin{lemma}\label{lem:gaussian_exp_tail}
Let $s^2$ be the sum of the diagonal elements of $\mSigma$. 
Assume $\E{Z}{\|\mathrm{NN}_{\vw}(Z)\|_2}<\infty$. 
For the density function $\ptheta(\vx)$ defined in \eqref{eq:gaussian_model}, for all $t>2s$ , we have    
\[
    P(\left|\|X\|-\mathbb{E}\|X\|\right| \geq t)\leq 2(e^{-\frac{t^2}{8L_1^2}} + e^{-\frac{(t/2-s)^2}{2L_2^2}})
\]
\end{lemma}
\begin{proof}
Note that 
\[
   P(\left|\|X\|-\mathbb{E}\|X\|\right| \geq t) \leq 
      P(\|X\|-\mathbb{E}\|X\| \geq t) + P(-\|X\|+\mathbb{E}\|X\| \geq t). 
\]
We bound the first term below (the second term can be handled similarly). 

We have    
\begin{align*}
P(\|X\|_2-E\|X\|_2\geq t) &= \E{Z}{P(\|X\|_2-E\|X\|_2\geq t\big|Z)}\\
&\leq \E{Z}{P(\|X\|_2-\E{X|Z}{\|X\|_2}\geq t/2\big|Z)} +  P(\E{X|Z}{\|X\|_2}-\E{}{\|X\|_2}\geq t/2).\\
\end{align*}
By Theorem \ref{thm:gaussian_concentration}, as $\ptheta(\vx|\vz)=\gN(\mathrm{NN}_\vw(\vz), \mSigma)$, 
\[
P(|\|X\|_2-\E{X|Z}{\|X\|_2}|\geq t/2\big|Z) \leq e^{-\frac{t^2}{8L_1^2}},
\]
where $L_1=\|\mSigma^{1/2}\|_{\mathrm{op}}$ is the operator norm of $\mSigma^{1/2}$. 
Therefore, 
\[
\E{Z}{P(|\|X\|_2-\E{X|Z}{\|X\|_2}|\geq t/2\big|Z)} \leq e^{-\frac{t^2}{8L_1^2}}.
\]
Let $\vmu(\vz)=\mathrm{NN}_\vw(\vz)$. 
For the second term, as 
\begin{align*}
&\E{X|Z}{\|X\|_2} \leq \sqrt{\E{X|Z}{\|X-\vmu(Z)\|_2^2}}+\|\vmu(Z)\|_2 = s + \lVert\vmu(Z)\rVert_2,\\
&\E{Z}{\|\vmu(Z)\|_2} = \E{Z}{\|\E{X|Z}{X}\|_2}\leq \E{Z}{\E{X|Z}{\|X\|_2}}=\E{X}{\|X\|_2}, 
\end{align*}
we have 
\begin{align*}
    P(\E{X|Z}{\|X\|_2}-\E{}{\|X\|_2}\geq t/2) \leq P(\|\vmu(Z)\|-\E{Z}{\|\vmu(Z)\|}\geq t/2-s). 
\end{align*}
By the Lipschitzness of $\mathrm{NN}_{\vw}(\vz)$ and $\ptheta(\vz)=\gN(0,\mathbf{I})$, we have for all $t>2s$ 
\[
P(\|\vmu(Z)\|-\E{Z}{\|\vmu(Z)\|}\geq t/2-s)\leq e^{-\frac{(t/2-s)^2}{2L_2}},
\]
where $L_2$ is the Lipschitz constant of $\mathrm{NN}_{\vw}$ (as a function of $\vz$). 
Combining these bounds gives
\[
P(\|X\|_2-E\|X\|_2\geq t) \leq e^{-\frac{t^2}{8L_1^2}} + e^{-\frac{(t/2-s)^2}{2L_2^2}}
\]
\end{proof}
\section{Method details}

\subsection{Alternative gradient models}\label{sec:other_gms}

To ensure that the estimate of $\Jtheta$ can  
be differentiated w.r.t.\ $\vtheta$ to obtain an estimate of $\Dxt$,
the gradient model needs to depend on model parameter $\vtheta$. 
KRR satisfies this condition in an attractive way, because its prediction depends on $\vtheta$ and $\vgamma$ in two separate factors, 
see \eqref{eq:krr_J}. 
However, though theoretically consistent, KRR estimates the gradient at the cost of $N^3$ in memory 
and time, where $N$ is the number of sleep samples. 
We discuss two alternative gradient models that could potentially be much faster, 
but there is no theoretical guarantee that $\ntheta\Jtr\evalt$ is close  to $\Dxt$. 

\subsubsection{Generic function approximator}
One can train a generic function estimator, such as a neural network, to estimate $\Jtheta(\vx)$. 
For such parametric models, the dependence on generative model parameters $\vtheta$ can 
be 
encapsulated into gradient model parameters $\vgamma$ through gradient descent. 
\begin{equation*}
   \vgamma(\vtheta) \leftarrow \vgamma(\vtheta) - \alpha \nabla_{\vgamma} L(\vtheta, \vgamma),\quad
   L(\vtheta,\vgamma) = \sum_{n=1}^N |\hat{J}_{\vgamma(\vtheta)}(\vx_n) - \lptheta(\vz_n,\vx_n)|^2
\end{equation*}
where $\alpha$ is the learning rate. 
As such, the estimator of $\Jtheta$ is better denoted as $\hat{J}_{\vgamma(\vtheta)}$ 
for a neural network with fixed hyperparameters.
Evaluating $\ntheta\hat{J}_{\vgamma(\vtheta)} \evalt$ can be implemented, 
though less straightforwardly compared to the KRR gradient model.
Alternatively, we can consider small perturbations around fixed-point of the 
loss, and derive 
a relationship between $\vgamma$ and $\vtheta$ at a local minimum:
\begin{align*}
    0 &= \frac{\partial L}{\partial \vgamma} (\vtheta+\ud\vtheta, \vgamma(\vtheta+\ud\vtheta))
    = \frac{\partial L}{\partial \vgamma} (\vtheta, \vgamma(\vtheta)) + 
      \ud\vtheta \frac{\partial}{\partial \vtheta} \frac{\partial L}{\partial \vgamma} (\vtheta, \vgamma(\vtheta)) +
      \ud\vgamma \frac{\partial}{\partial \vgamma} \frac{\partial L}{\partial \vgamma} (\vtheta, \vgamma(\vtheta)).
\end{align*}
The first term on the RHS is zero, and rearranging gives 
$\frac{\ud\vgamma(\vtheta)}{\ud\vtheta}
=-\left(\frac{\partial^2 L}{\partial \vgamma\partial\vgamma}\right)^{-1}\frac{\partial^2 L}{\partial\vtheta \partial\vgamma}$, assuming the inverse exists.
Thus,
\begin{equation*}
    \frac{\ud\hat{J}_{\vgamma(\vtheta)}(\vx)}{\ud\vtheta} 
    = \frac{\partial \hat{J}_{\vgamma(\vtheta)}(\vx)}{\partial \vgamma}\frac{\ud \vgamma(\vtheta)}{\ud\vtheta}
    = -\frac{\partial \hat{J}_{\vgamma(\vtheta)}(\vx)}{\partial \vgamma}
        \left(\frac{\partial^2 L}{\partial \vgamma\partial\vgamma}\right)^{-1}\frac{\partial^2 L}{\partial\vtheta \partial\vgamma}.
\end{equation*}
All of the factors can be computed by automatic differentiation since the objects being differentiated 
are all scalars. However, for a generic neural network, the Hessian of the loss 
w.r.t.\ $\vgamma$ may not exist, and computing it can be unstable.

\subsubsection{Particle estimator} \label{sec:particle_details}
The prediction of the KRR estimator may not 
but a valid expectation. In other words, $\Jtr(\vx)$ may not correspond 
to the expected log joint under any valid probability distribution.
To address this issue, we can approximate $\Dxt$ through a set of particles $\vz'$ (in the space of the latent)  
generated from a simulator $S_\vgamma:(\vx,\vn)\to\vz'$, where $\vgamma$
is the parameter of the simulator, $\vx$ is an observation, and $\vn$ is a noise 
source distributed as $\zeta(\vn)$.
For all $\vx$ from the generative model, we want the simulator to produce particles 
such that 
$\E{\zeta(\vn)}{\ntheta \log \ptheta(S(\vx,\vn),\vx)}\evalt$ estimates of $\Dxt$.
This can be achieved by solving
$$
\min_\vgamma \E{\pthetat(\vz,\vx)}{ \left \| \E{p(\vn)}{\ntheta \log \ptheta(S_\vgamma(\vx,\vn),\vx)}\evalt  - \nlptheta(\vz,\vx)\evalt\right\|^2},
$$
which is equivalent to
$$
\min_\vgamma \E{\pthetat(\vx)}{ \left \| \E{p(\vn)}{\ntheta \log \ptheta(S_\vgamma(\vx,\vn),\vx)}\evalt  
- \E{\pthetat(\vz|\vx)}{\nlptheta(\vz,\vx)\evalt}\right\|^2}
$$
due to the property of mean squared error (see \cref{sec:mse_for_mean_proof}). 
We know that the optimal set of particles is distributed as the posterior $\pthetat(\vz|\vx)$, 
but minimising the cost above
does not necessarily drive $S_\vgamma$ to produce posterior samples. Nonetheless, this set of 
particles is adequate to approximate $\Dxt$. 
We refer to this scheme as amortised learning by particles (AL-P). 
We test this on sample quality experiments and found that the KIDs and FIDs
were in general worse than even the vanilla VAE. 
Samples from the model trained by AL-P are shown in \cref{fig:real_mnist_0_conv} to \cref{fig:real_celeb_0_conv} 
in section \cref{sec:real_details}.

\subsubsection{Relationship between KRR gradient model and importance sampling}
The KRR gradient model approximates $\Dxt$ by 
linearly weighting $\{\nlptheta(\vz_n,\vx_n)\}_{n=1}^N$.
This is similar to other reweighting schemes (e.g.\ \citep{DiengPaisley2019Reweighted}), with the most simple one being
importance sampling where the proposals are from the prior $\ptheta(\vz)$, 
and the weights are normalised density ratios $\ptheta(\vz,\vx)/\ptheta(\vz)$.
Importance sampling is an unbiased estimation method, but has huge variance and 
requires at least exponentially
many samples as the KL divergence between the posterior and prior \citep{ChatterjeeEtAl2018}.

It would then appear that KRR should perform similarly with importance sampling
in estimating $\Delta(\vx)$, but, on closer look, they use slightly different 
sources of information for estimation.
KRR uses a set of samples $(\vz_n, \vx_n) \sim \ptheta(\vz,\vx)$, whereas importance 
sampling uses $\vz_n \sim \ptheta(\vz)$ and $\ptheta(\vz,\vx^*)$.
In computing the weights for a particular $\vx^{*}$ from the dataset, 
KRR compares $\vx^{*}$ with all sleep samples $\{\vx_n\}_{n=1}^N$, 
using a similarity metric  
determined by the kernel function. The weights $\valpha$ also takes into account of the similarities between 
all sleep samples. 
On the other hand, importance sampling uses 
$\ptheta(\vz,\vx^{*})$ for a given $\vx^*$ and computes the weights 
for each sample of $\vz$ independently of each other. 
In addition, the importance sampling weights are constrained to be 
non-negative and sum up to one, whereas the weights in KRR are not constrained and thus can be more flexible.

\subsection{Kernel architecture}\label{sec:kernel_details}
In all experiments, we used a squared-exponential kernel 
$k(\vx,\vx')=\exp(-0.5\|\vphi_\vw(\vx)-\vphi_\vw(\vx')\|_2^2/\sigma^2)$.
The feature $\vphi_\vw$ can be the identity function, a linear projection, or a linear projection followed by batch normalisation 
, see \cref{tabel:params} which lists the architectures used for each experiment.
The linear projection and batch normalisation are primarily used on high-dimensional benchmark datasets.
Nonlinear projections, such as 
deep neural networks, did not give significant improvement while consuming more 
memory.
The bandwidth $\sigma$ is initialised as the median of the distance 
between $\vphi_\vw(\vx^{(n)})$ where $\vx^{(n)} \sim \pthetat(\vx)$. 

\begin{sidewaystable}[!htbp]
\centering
\begin{tabular}{l|*{11}{c}}
    Experiment              & latent dim    & data dim   & $M$(data) & $N$ (sleep)  & $L$ (val) &  $\lambda$  & \# proj  & batch norm?   & gen lr &  grad lr  &  nepoch   \\
    \hline                  
    gradient estimation     & 2             & 1          & 100       & 5\,000         & --        &  0.01(f)   &  --        &  no           & --        &  --       &  --       \\
    spherical prior         & 1             & 256        & 10\,000    & 2\,000        & 200       &  0.01       &  300       &  no           & 0.001     &  0.001    & 30        \\   
    pinwheel                & 2             & 2          & 2\,500     & 1\,000        & 200       &  0.01       & --         &  no           & 0.001     &  0.001    & 2\,500      \\
    Independent component   & 36            & 256        & 100\,000   & 2\,000        & 200       &  0.001(f)  &  300       &  no            & 0.001     &  0.01     & 100       \\
    Matrix factorisation    & 100           & 784        & 5\,000     & 2\,000          & 150       &  0.001      &  300       &  yes          & 0.001     &  0.001    & 300       \\
    neural process          & 50            & 8              & 10\,000    & 4\,000        & 200       &  0.001      &  --        &  no           & 0.0001    &  0.0001   & 50        \\
    nonlinear oscillation   & 2/time step   & 600        & 1         & 2\,000        & 200       &  0.001      &  200       &  no           & 0.001     &  0.001    & 5\,000  \\
    Hodgkin-Huxley          & 3/time step   & 1\,000      & 1         & 2\,000        & 200       &  0.001      &  200       &  no           & 0.001     &  0.001    & 50\,000 \\
    ecology                 & 1/time step   & 180        & 1         & 2\,000        & 200       &  0.001      &  --        &  no           & 0.001     &  0.001    & 50\,000 \\
    B-MNIST                 & 16            & 1\,024      & 60\,000    & 2\,000        & 200       &  0.1 (f)    &  300      &  yes        & 0.001     &  0.001    & 50 \\
    MNIST                   & 16            & 1\,024      & 60\,000    & 2\,000        & 200       &  0.1 (f)    &  300      &  yes        & 0.001     &  0.001    & 50 \\
    Fashion                 & 16            & 1\,024      & 60\,000    & 2\,000        & 200       &  0.1 (f)    &  300      &  yes        & 0.001     &  0.001    & 50 \\
    Natural                 & 16            & 1\,024      & 100\,000   & 2\,000        & 200       &  0.1 (f)    &  300      &  yes        & 0.001     &  0.001    & 50 \\
    CIFAR                   & 16            & 3\,072      & 50\,000    & 2\,000        & 200       &  0.1 (f)    &  300      &  yes        & 0.001     &  0.001    & 50 \\
    CelebA                  & 16            & 3\,072      & 100\,000   & 2\,000        & 200       &  0.1 (f)    &  300      &  yes        & 0.001     &  0.001    & 50 \\
    \hline
\end{tabular}
\caption{Data properties, and model and training parameters of ALWS for each experiment.
The regularisation strength $\lambda$ is sometimes fixed as indicated by (f). 
See \cref{sec:kernel_details} for kernel architectures.
}
\label{tabel:params}
\end{sidewaystable}

\section{Experimental details}\label{sec:exp_details}
We list the model and training parameters used to run each experiment in \cref{tabel:params}.
The batch size is 100 except for dynamical models and neural process where the batch size is 1.

\subsection{Gradient estimation}\label{sec:gradient_details}

The toy generative model has 
$\evz_1,\evz_2\sim \gN(0,1), x|z\sim\gN(\softplus(\vw \cdot \vz)-\|\vw\|_2^2, \sigma_2^2)$.
The observations are 100 samples for drawn form the model with 
$w_1=w_2=1, \sigma=0.1$. Note that the ML solution for this 
synthetic problem is not unique.

For variational learning,
the approximate posterior is a factorised Gaussian
that minimises the ELBO. 
The gradient of ELBO
was approximated by samples. 
The mean
and variances are initialised as the standard Gaussian and 
are optimised by Adam with step size 0.01 for 300 iterations, which
is sufficient for convergence. For ground truth, we estimated the 
gradient by importance sampling, with $5\times 10^4$ samples proposed from the prior.

\subsection{Spherical prior}

The data are $16\times 16$ Gabor images. The orientation is uniformly distributed 
over one period $0$ to $\pi$.
The generative network is taken from the first two deconvolutional layers of DCGAN 
so that the output size is $16\times 16$. 
For VAE, we used the symmetric convolutional neural network for the encoder 
and a factorised Gaussian posterior.
For $\mathcal{S}$-VAE, a von Mises-Fisher distribution is used as the posterior.

\subsection{Hierarchical models}\label{sec:pinwheel_details}

The penalty assigned to probability vector $\vm$ in the categorical distribution 
is the log pdf of a Dirichlet prior 
$\log p(\vq)=(\alpha-1)\sum_i \log q_i + \text{const}$, where $q_i=e^{m_i}/\sum_j e^{m_j}$.
We use $\alpha=0.999$. 
Similarly, for the $k$'th component
in the mixture, the Normal-InverseWishart distribution
has log-likelihood that penalises $\|\mu_k\|$, $\log |\Sigma_k|$ and $\textrm{Tr}(\Sigma_k^{-1})$. 
In addition, we also penalise the $l$-2 norm of neural network weights. These penalisation 
strengths are set to $10^{-4}$. 

The relative maximum mean discrepancy (MMD) test \citep{BounliphoneGretton2016Test} is used for model comparison
based on generated samples. 
Denote the set of real data by $\gD$ and the set of generated samples from model A by 
$\gD^\textrm{A}$. The null hypothesis for this test is 
$\mathrm{MMD}(\gD,\gD^\textrm{SIN}) < \mathrm{MMD}(\gD,\gD^\textrm{ALWS})$,
where $\mathrm{MMD}$ is the MMD distance between two sets of samples.
The test returns a $p$-value of 0.514 based on 1500 samples from each of the 
three distributions, 
suggesting that the two models perform almost equally
well on learning this data distribution. 
We note that SIN is trained on a full Bayesian version
of the model, and the samples are reconstructions given the real dataset, 
giving an advantage for SIN.

\subsection{Parameter identification} \label{sec:ica_details}

\begin{figure}
    \centering
    \includegraphics[width=0.5\textwidth]{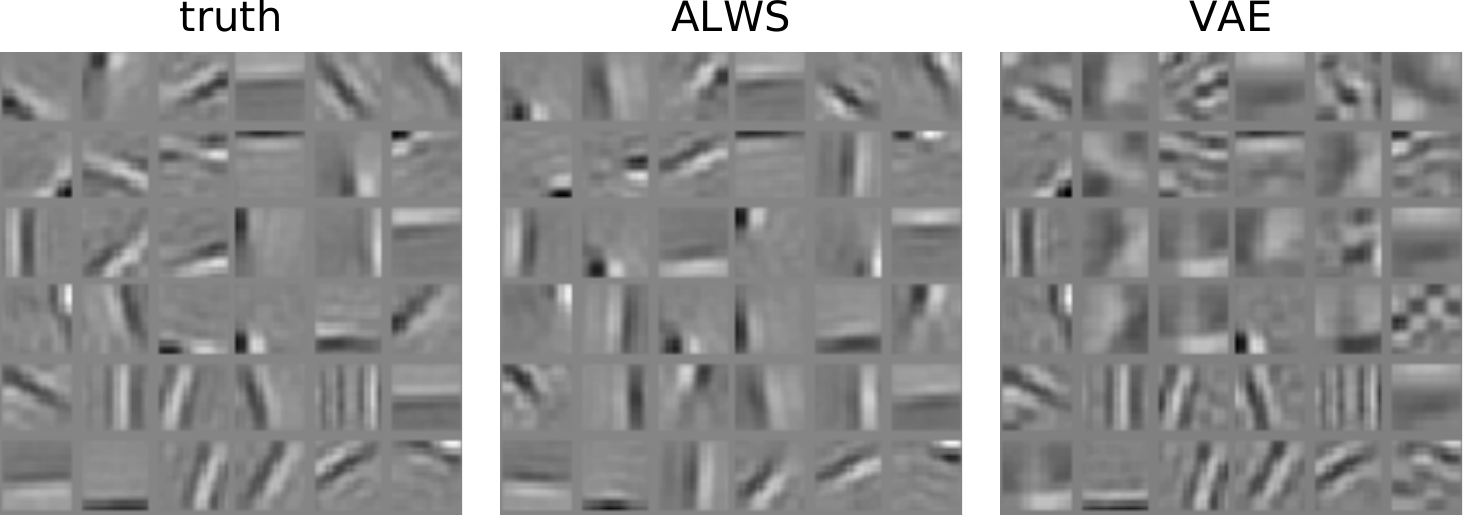}
    \caption{Same as \cref{fig:ica} but with $\lambda$ adaptive.}
    \label{fig:ica_adaptive}
\end{figure}

The linear basis (weights) are the top 36 independent components 
of natural images discovered by the FastICA algorithm. 
Each component is subtracted by their mean and normalised to have 
unit length. The synthesised dataset is standardised 
by subtracting the mean and dividing by the standard deviation. 
The kernel is augmented with an adaptive linear neural network feature with 
300 outputs. Using 200 features produces very similar results. 
The regularisation strength is fixed at  $\lambda=0.001$. 
Adapting the filters results in slightly different filters as shown in
\cref{fig:ica_adaptive}. 

\subsection{Neural process}\label{sec:np_details}

\paragraph{Introduction.}
We briefly review the neural processes (NPs, \citet{garnelo2018neural}). 
Suppose there is a distribution over function $f\sim \mathcal{P}(f),\ f: \gX\to\gY$.
We observe information a given function $f$ through its potentially noisy values at a set of inputs $(\vx, \vy)|f$.
The task is the following: given a set of context pairs $\gD:=\{(\vx^C_k,\vy^C_k)\}_{k=1}^K$ drawn from an unobserved function, 
infer the distribution of the function value at a set of target inputs $\{\vx^T_m\}_{m=1}^M$. 

NPs represent the posterior of $f$ given $\gC$ by a random variable $\vz$, 
which is combined with $\vx_m^T$ to predict the function value.
During training, the training data comprises multiple sets of input-output pairs, and each set is always 
conditioned on one particular $f\sim \gP$. 
The training data are split into a context set $\gC$, 
used to condition the representation $\vz$, 
and a target set $\{(\vx^T_m, \vy_m^T)\}_{m=1}^M$, used to evaluate  
the likelihood of $\vy_m^T$ given $\vz$ and $\vx_m^T$.
Formally, the generative model is specified by
\begin{align*}
    \vr &= \frac{1}{K} \sum_{k=1}^K \vrho_\vtheta(\vx^C_k, \vy^C_k) \\
    p_\vtheta(\vz|\vr) &= \gN(\vz|\vmu^C_\vtheta(\vr), \mSigma^C_\vtheta(\vr))\\
    p_\vtheta(\{\vy_m^T\}|\{\vx_m^T\},\vz) &= \prod_{m=1}^M \gN(\vy_m^T|\vmu^T_\vtheta(\vz,\vx_m^T), \mSigma^T_\vtheta(\vz,\vx_m^T)).
\end{align*}
In short, a latent representation of the context $\vz$ is drawn from a normal 
distribution with parameters formed by an exchangeable function of the context set $\gC$, 
and the likelihood on the target outputs are i.i.d. Gaussian conditioned on $\vz$ and $\vx_m^T$. 
The objective for learning is  
to maximise the likelihood of the target output conditioned on the corresponding 
context set from the same underlying $f$ and the target input. Once trained,
 the neural process is able to produce samples from the distribution of function values (target outputs) 
at context inputs.

The encoding function $\vrho_\vtheta$ 
plays the role of an inferential model, but we can view it 
as a function that parametrises the ``prior'' distribution on $\vz$ given the context set,
and the parameters in $\vrho$ can be regarded as belonging to the generative model. The gradient model
trained by KRR also needs to be conditioned on each context set, but for simplicity,
we train a gradient model for a single context followed by $\vtheta$ update.
\citet{garnelo2018neural} trained the neural processes by maximising an ELBO with posteriors 
of the form 
$$q(\vz|\gC, \gT)=
\ptheta(\vz|\vr^{CT}), \quad
\vr^{CT} = \frac{1}{K} \sum_{k=1}^K \vrho_\vtheta(\vx^C_k, \vy^C_k) +\frac{1}{M} \sum_{m=1}^M \vrho_\vtheta(\vx^T_m, \vy^T_m),
$$
which is an approximation.

\begin{figure}[t]
    \centering
    \includegraphics[width = 0.4\textwidth]{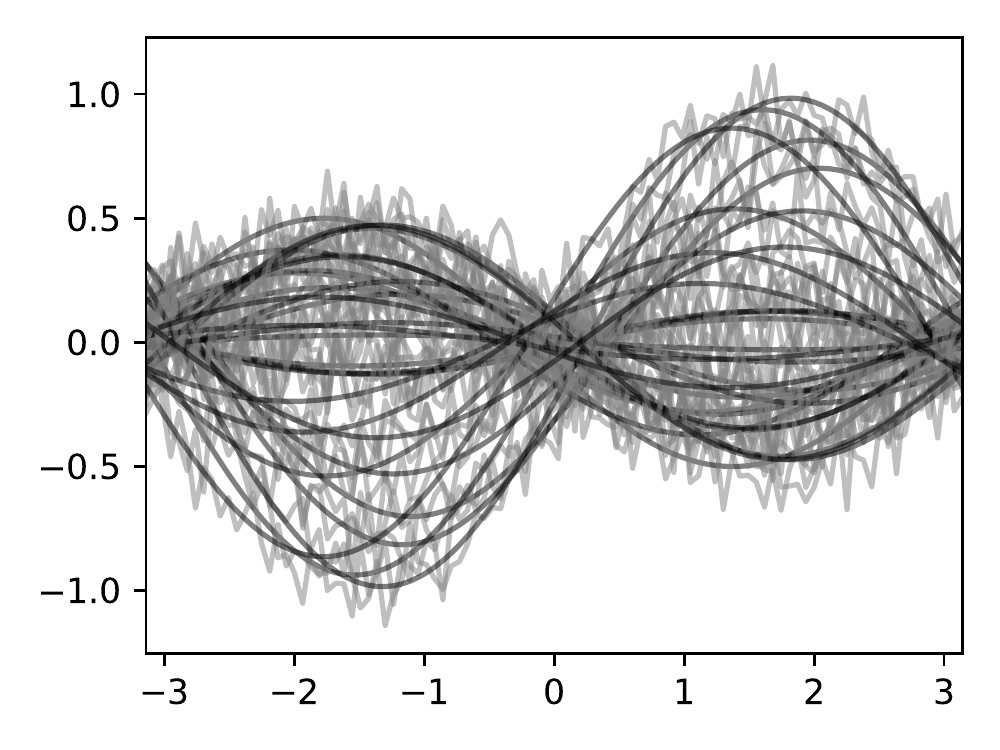}\\
    \includegraphics[width=1.0\textwidth]{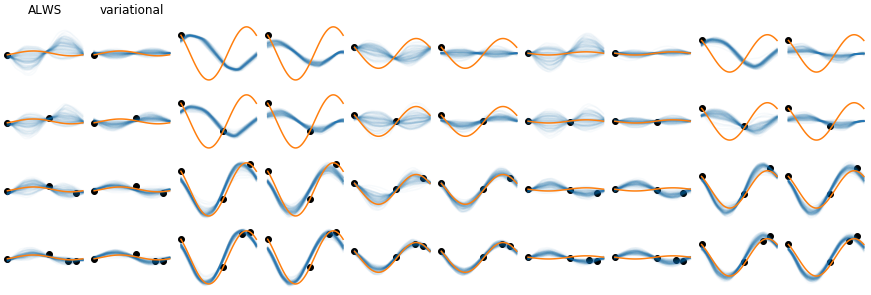}
    \caption{Neural processes. 
    Top: samples from prior distribution of functions. Black: Latent function. Grey: noisy observations.
    Bottom: posterior samples (blue lines) from ALWS (odd columns) and the original variational method (even columns).
    Orange lines are true latent functions $f$. Black dots are context pairs.
    }
    \label{fig:np}
\end{figure}

\paragraph{Experiments.} 
We train a neural process on a $\gP(f)$ that have samples as shown in \cref{fig:np} (top).
They are sinusoids with random amplitudes and phase shifts and supported on $[-\pi, \pi]$. 
The observations are contaminated with Gaussian noise with 
standard deviation 0.1.
Conditioning the function with a context input around $-\pi$, $0.0$ and 
$\pi$ induces large uncertainty
over $f$; thus, we can use this to probe the representation of uncertainty.

In the NP model, the representation $\vr$ and $\vz$ 
are both 50-dimensional. And the encoding and decoding networks 
are fully connected with ReLU nonlinearities.
During training, the number of context pairs $K=4$, and the target set contains the context pairs and an additional four pairs, 
so $M=8$ and $\gC\subset\gT$.
The gradient model is trained for each given context set, and hence the 
batch size is 1.
A small learning rate of 0.0001 is used for all models and parameters.
The gradient model is trained to take sleep samples $\vy_m^T$ evaluated for this single $\gC$ at each $\vx_m^T$.
The kernel takes $\{\vy_m\}_{m=1}^8$ as a single vector.
We note that other kernels on sets could be used.

During test time, we evaluate the predicted function value of a dense grid of points in 
$[-\pi,\pi]$ given 1 to 4 context pairs. As shown in \cref{fig:np} (lower panels), when the 
number of context points is small,
the model trained with ALWS makes more accurate predictions, 
and better reflects the uncertainty of the function value when the context set is uninformative.
Given four context pairs (as in training),
we test the learned model on 500 functions from $\gP(f)$ and evaluate 
how close samples of $\gP(\vf|\gC,\vx_m^T)$ are to the true function at 
$M=100$ target locations. 
We use either the posterior mean or 
a random posterior sample from the posterior as a point estimate,
and measure the performance by mean squared error.
We find that the errors are significantly smaller for ALWS-trained model based on paired tests 
for the posterior mean prediction 
(paired t-test, $t=-3.47, p=0.00056$; mean of ALWS, -5.11; variational, -4.99. 
Wilcoxson test, $W=44837.0, p=3.7\times10^{-8}$, median of ALWS, -5.11, variational, -4.98)
and the random sample prediction
(paired t-test, $t=-2.09, p=0.037$; mean of ALWS, -4.87; variational, -4.77. 
Wilcoxson test, $W=53762.0, p=0.0061$, median of ALWS, -4.91, variational, -4.69).

\subsection{Nonlinear dynamic model}\label{sec:dynamic_details}

We run ALWS for generative models whose priors are defined through 
nonlinear transitions in time.
In all of the experiments, 
we treat each sequence as a single multi-dimensional data point. 

\begin{figure}
    \centering
    \includegraphics[width=0.5\textwidth]{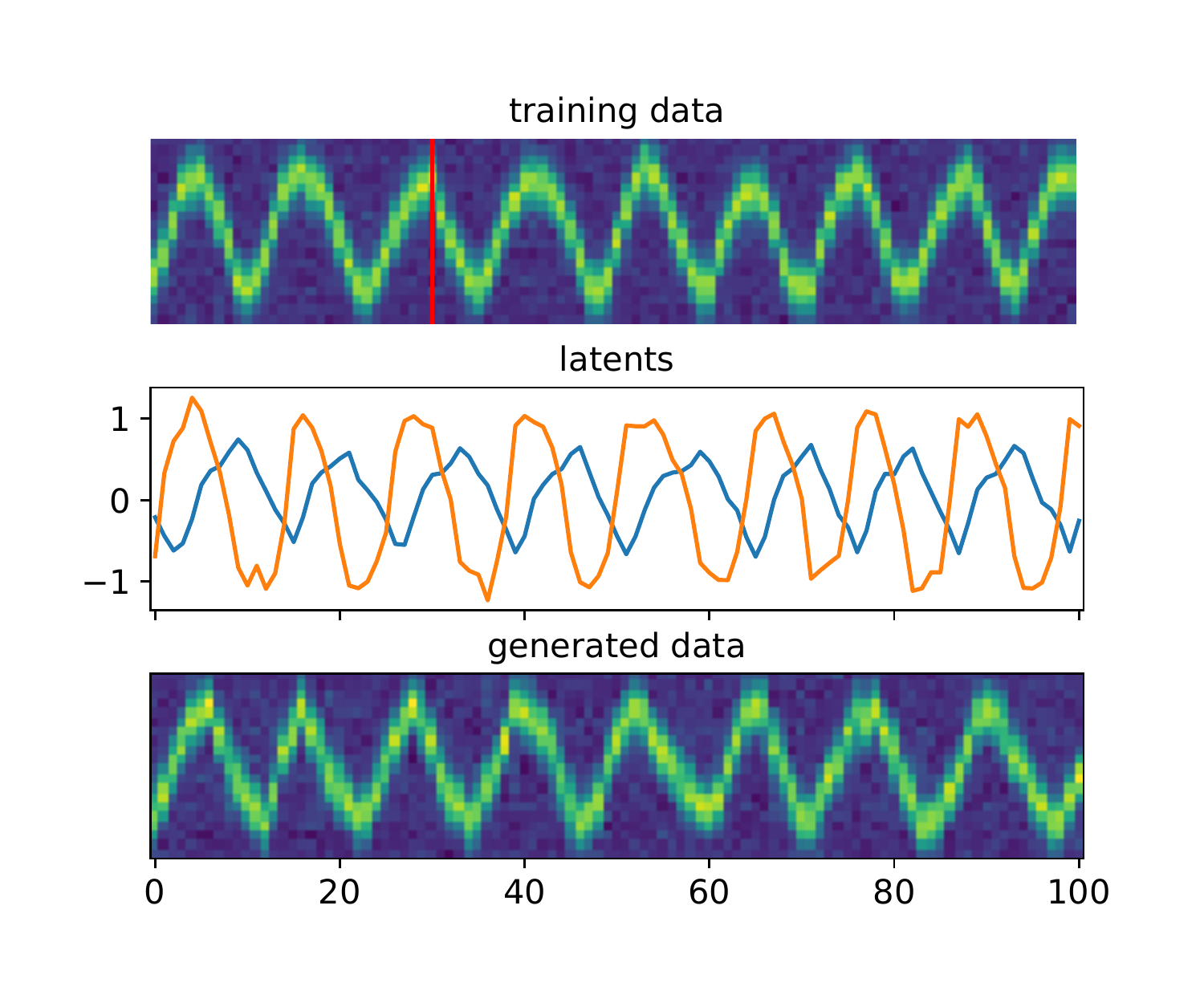}
    \caption{Nonlinear oscillations. 
    Top: an example trajectory. Only the first 30 time steps marked by the red line 
    is used for training. Three such 30 time step traces are used for training.
    Middle: latent space learned by ALWS.
    Bottom: Generated trajectory.}
    \label{fig:oscillations}
\end{figure}

\subsubsection{Nonlinear oscillations}
We generate data from a nonlinear oscillation process according to the following 
equations used by \citet{WenliangSahani2019neurally}
\begin{gather*}
    \vz_t = \operatorname{Rot}(\vz_{t-1})+\vepsilon_t^{(z)},    \quad    \vx_t = \operatorname{Img}(z_{t,1}) + \vepsilon_t^{(x)}\\
    \operatorname{Rot}(\vz_t) = \mR_\alpha \vz_{t}\frac{r(\|\vz_t\|_2)}{\|\vz_t\|_2}, 
    \quad r(a)= \operatorname{sigmoid}(4 ( a-0.3) ), 
    \quad [\operatorname{Img}(z)]_i = \exp(-0.5(z-\bar{z}_i)^2/0.3^2)
\end{gather*}
where $\mR_\alpha$ is a rotation matrix by $\alpha$ radians, $\operatorname{Img}$ maps one of the latent dimensions into a 20-pixel image through
Gaussian bumps with evenly spaced centers at $\bar{z}_i, i\in\{1,\dots,20\}$. 
Intuitively, the latent $\vz$ is rotated by $\alpha$ and scaled radially so that its length remains close to 1. 
Samples of $\vx_t$ for all $t\in\{1,\dots,T\}$ can be 
plotted side by side as 
a $20\times T$ image, which is shown in \cref{fig:oscillations} (top).

We train the following generative model:
\begin{gather*}
    \ptheta(\vz_t|\vz_{t-1}) = \gN(\vz_t;\text{NN}_\vw^{(z)}(\vz_{t-1}), \mSigma_z),\qquad
    \ptheta(\vx_t|\vz_{t}) = \gN(\vx_t;\text{NN}_\vw^{(x)}(\vz_{t}), \mSigma_x),
\end{gather*}
where the parameters are the weights and biases in the neural networks (NN), and 
the diagonal covariance matrices $\mSigma_{(\cdot)}$'s.
The number of units 
are fully connected with $2\to 20 \to 2$ neurons for $\text{NN}^{(z)}$ and 
$2\to 20 \to 20$ for $\text{NN}^{(x)}$. The $\tanh$ is used as the nonlinearity.
We train the model on a single sequence of 30 time steps and then generate a 100-step
 sequence of the learnt latents and observations shown in \cref{fig:oscillations}. The latents
 correctly capture the position, which directly sets the data, and the velocity, 
 which needs to be learned from data.

\subsubsection{Hodgkin-Huxley (HH) equations}
The HH equations are described by
\begin{align*}
    C_m\dot{V}(t) &= -g_l[V(t) - E_l]
                            - \bar{g}_N m^3(t) h(t) [V(t) - E_N] 
                            - \bar{g}_K n^4(t) [V(t) - E_K] 
                            + I_{\text{in}}(t) +\epsilon(t)\\
    \dot{e}(t)     &=  \alpha_e(V(t)) [1 - e(t)] -\beta_e(V(t)) e(t), \quad e\in\{m, h, n\}
\end{align*}
where $\alpha_e$ and $\beta_e$ are nonlinear functions of $V(t)$ involving a parameter $V_T$ that sets the threshold 
for action potentials, see \cite{PospischilDestexhe2008Minimal} for details.

We used forward-Euler method for simulation with a time step of $\Delta t=0.05ms$. 
At each step of the simulation, we add a small Gaussian noise of standard deviation $\sigma_z = 0.1$mV to $V_t$ as process noise.
The measurements noise added to observations (but not propagated to $V_{t+1}$) 
is Gaussian with standard deviation $1.0$mV.
There 10 parameters for the resulting discrete-time state-space model: 
$\vtheta=\{C_m, g_l, E_l, \bar{g}_N, E_n, \bar{g}_K, E_K, V_t, \sigma_z, \sigma_x\}$.

\begin{figure}
    \centering
    \includegraphics[width=0.75\textwidth]{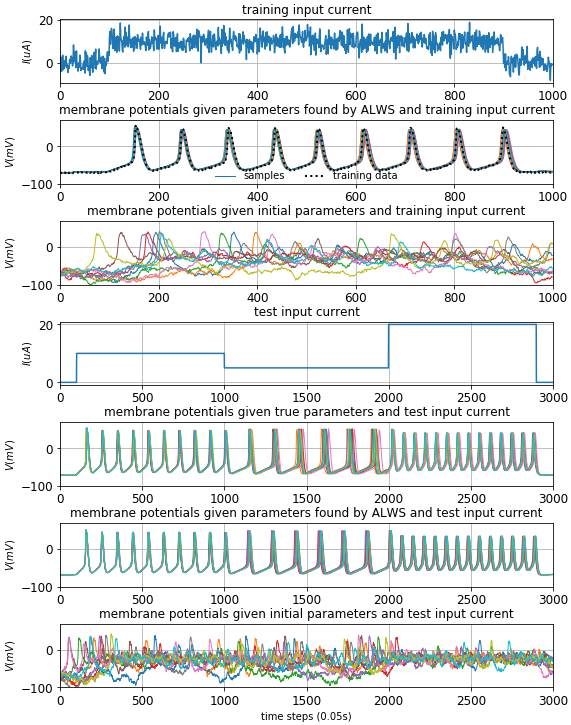}
    \includegraphics[width=0.8\textwidth]{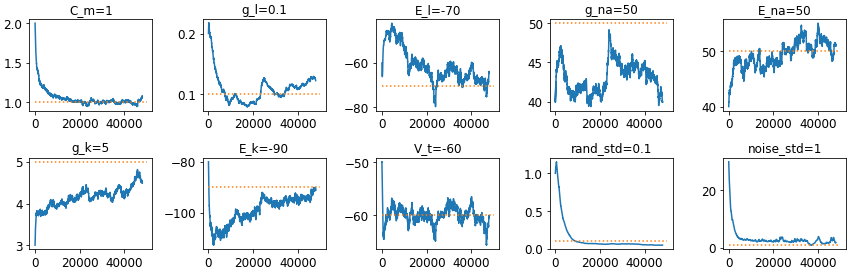}
    \caption{Hodgkin Huxley simulations. 
    Top seven panels:
    1st row, input current $I_{\text{in}}$ during training. 
    2nd-3rd rows, trajectories given learnt and initial parameters under
    training input current.
    4th row, test input current.
    5th-7rd rows, trajectories given true, learnt and initial parameters under
    test input current.
    Bottom 10 panels:
    Blue solid: parameter value at each iteration. 
    Yellow dashed: true parameter values.
    }
    \label{fig:hh}
\end{figure}

We train and test the model under different input current sequences $I_{\text{in}}$.
The results are shown in \cref{fig:hh}. 
We simulate a single trajectory from the model
with some true parameters 
and a noisy current injection shown in \cref{fig:hh}(1st row). 
This sequence is used as the training data \cref{fig:hh}(2nd row, dotted).
We then perturb these
parameters, making the simulated trajectories unrealistic 
\cref{fig:hh}(3rd row).
After training, the simulated trajectories look almost identical to the training
data \cref{fig:hh}(2nd row, solid).
To test whether the learned model can be used for prediction under a different current injection, 
we simulate trajectories given an unseen test current
\cref{fig:hh}(4th row).
The responses of membrane potential under true parameters are shown in 
\cref{fig:hh}(5th row). Samples from the trained model \cref{fig:hh}(6th row, solid)
under this unseen current  
are very similar to the trajectories given real parameters, showing 
generally correct phase, periodicity and amplitude. 
The simulated responses have less variation between trajectories, which could be 
due to training under a single sequence. Indeed, not all parameters converge
to the true parameters \cref{fig:hh} (bottom panels). 

\begin{figure}
    \centering
    \includegraphics[width=0.8\textwidth]{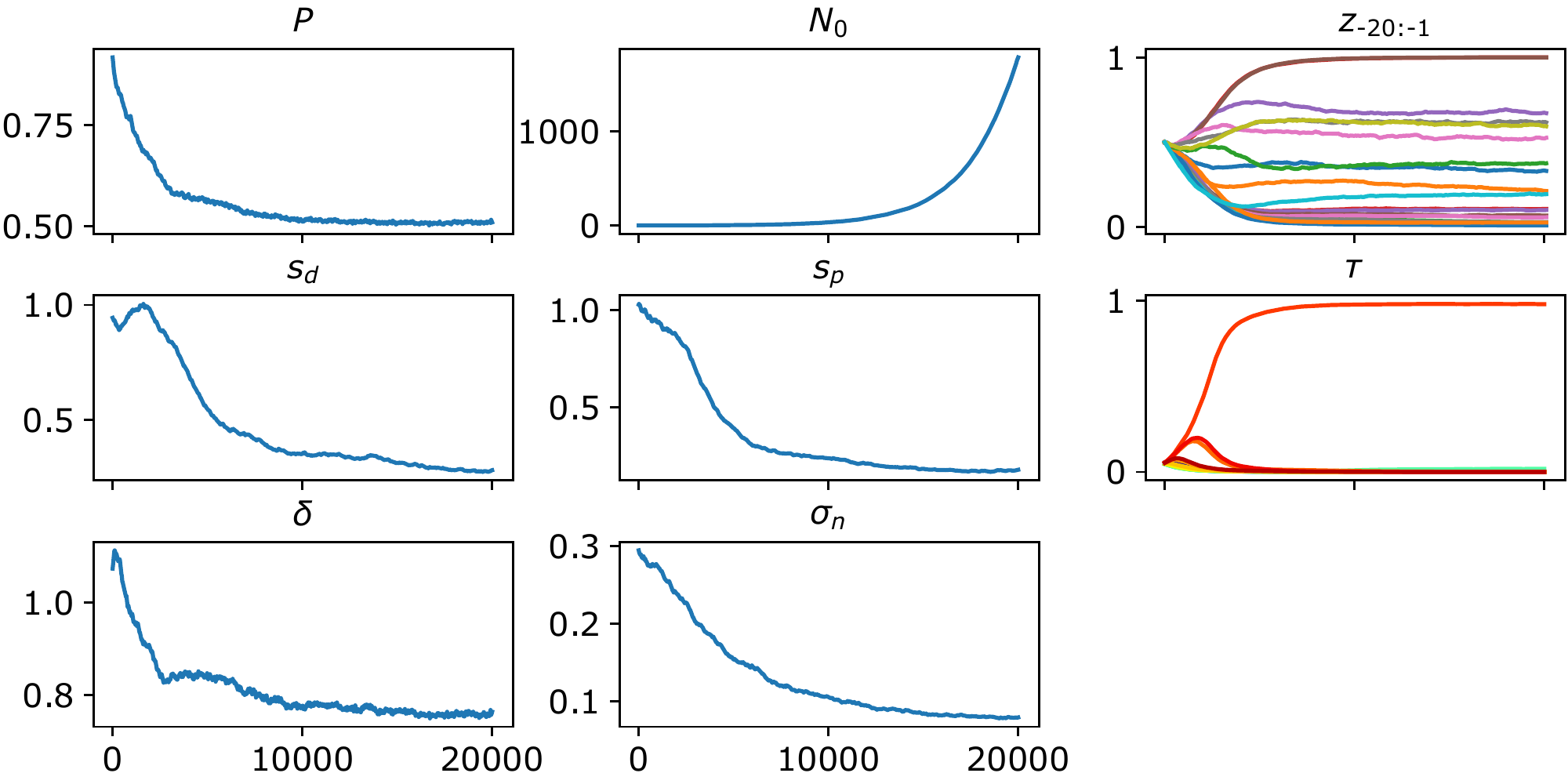}
    \caption{Evolution of parameters in the ecological model for blowfly population.}
    \label{fig:eco_details}
\end{figure}

\subsubsection{Ecological data}\label{sec:eco_details}
We train a model that describes the evolution of blowfly population size 
under food limitation \citep{WoodWood2010Statistical}.
The model is given by
\begin{gather*}
    \tau \sim \mathrm{Categorical}(\vm), \tau \in \{1,\dots,20\},\quad
    e_t \sim \textrm{Gamma}(\frac{1}{\sigma_p^2}, \sigma_p^2),\quad
    \epsilon_t \sim \textrm{Gamma}(\frac{1}{\sigma_d^2}, \sigma_d^2),\\
    z_t = Px_{t-\tau} \exp(-\frac{x_{t-\tau}}{N_0}) + x_t \exp(-\delta \epsilon_t),\quad 
    p(x_t|z_t) = \textrm{LogNormal}(\log(z_t),\sigma_n^2)
\end{gather*}
Note that $\tau$ is a discrete delay drawn from a categorical distribution with 
logit parameters $\vm$, $e_t$ and $\epsilon_t$ are stochastic 
variations in births and deaths 
following Gamma distribution with a common mean 1.0 and standard deviations $\sigma_p^2$ and $\sigma_d^2$, respectively.
The observation is noisy 
with log-normal noise so that $\vx_t$ remains positive. 
Observations in the first 20 time steps depend on some past data that is not observed, 
so we modelled these past data $x_{-20:-1}$ as parameters, 
which are constrained to be between 0 and 1.0. 
Thus, this model has parameters $\vtheta=\{\vm, \sigma_d, \sigma_p, P, N_0, \delta, \sigma_n, x_{-20:-1}\}$

We fit the model on a data sequence of length 180, normalised to be between 0 and 1.0. 
The evolution of parameters is shown in \cref{fig:eco_details}. 
As our training objective is different from that of ABC methods, 
we do not make direct quantitative comparison with them.
But compared with the samples from three ABC methods shown in 
\citep{ParkSejdinovic2016K2-ABC} (Figure 2B), 
it is clear that samples from ALWS are visually more similar to the training data.

\subsection{Sample quality on benchmark datasets}\label{sec:real_details}

\subsubsection{Data processing}
All images have $32\times32$ pixels by their original sizes (Natural, CIFAR-10), or by zero-padding (MNIST,F-MNIST) or interpolation (CelebA). 
The binarised MNIST is statically binarised once before training. Each pixel is set to 1 with probability 
equal to the pixel value after rescaling to between 0 and 1.
The natural images \footnote{\url{github.com/hunse/vanhateren}} are 
patches from large natural scenes. No clipping is applied.
Original MNIST Fashion MNIST, CIFAR-10 and CelebA images are rescaled to between $-1.0$ and $1.0$.

\begin{figure}[t]
    \centering
    \includegraphics[width=\textwidth]{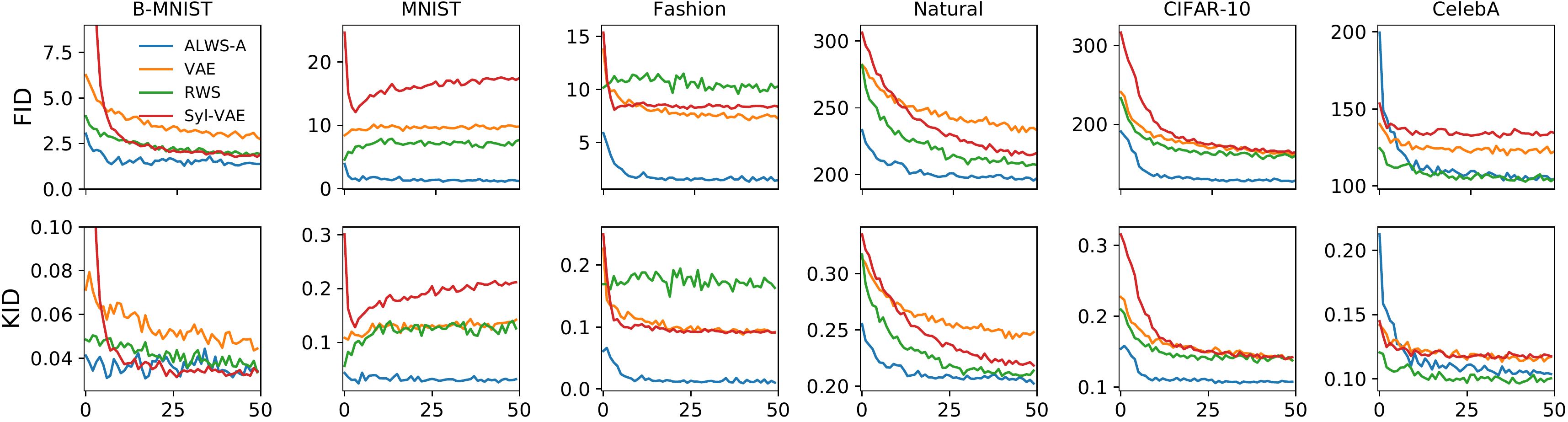}
    \includegraphics[width=\textwidth]{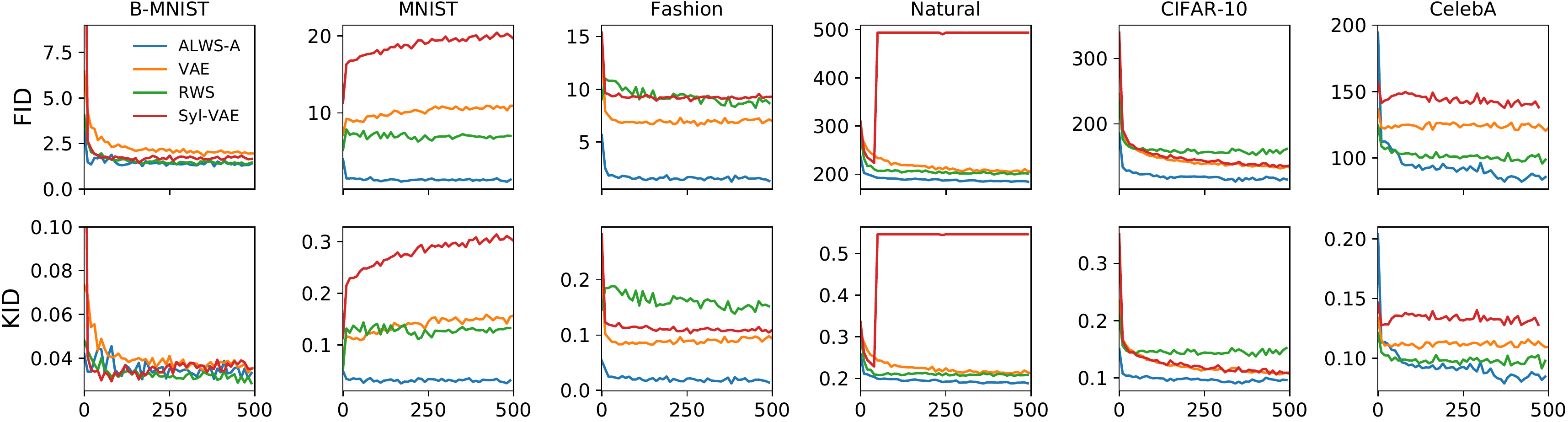}
    \caption{KID and FID scores at the end of each epoch for selected algorithms on convolutional architecture.  
    Top two rows show distances during a run of 50 iterations at every iteration, and 
    the bottom two rows show another run of 500 iterations at every 10th iteration.}
    \label{fig:real_kid_epoch}
\end{figure}

\subsubsection{Model and training details}
All methods use the same neural network as the DCGAN without the last convolutional 
layer to make the image size $32\time32$. Batch size is 100 for each 
update of generative and gradient model parameters. 
We run each algorithm on each dataset with 10 different initialisations.
The neural network in the generative model has $\operatorname{ReLU}$ nonlinearities 
in intermediate layers.
The nonlinearity for the final layer depends on the dataset: 
it is sigmoid for binary MNIST, linear natural images, 
and $\tanh$ for the other datasets. 

All methods are trained for 50 epochs except for SIVI which was trained for 1\,000 epochs.
The optimizer is Adam with a fixed learning rate of 0.001.
For ALWS, we use $2\,000$ sleep samples for training the gradient model.
The kernel is augmented by the linear projection to $300$ dimensions for all datasets. 
A larger number of output dimension produced better results but induces longer run time.
The weights of the projection are updated after the first five epochs. 
The regularisation parameter 
$\lambda$ is fixed at 0.1; this helps sample quality for CIFAR and CelebA, but does not affect
or worsens sample quality for the other datasets.
For ALWS-F, a fixed random projection is used throughout training. 
For ALWS-A, the linear weights are training at each parameter update 
after five epochs, using the two-stage training. 

For VAE, the encoder network is symmetrical to the 
generative network and is appended with a final linear layer for posterior statistics. 

For Syl-VAE. We change the gated convolutional layer in the decoder network
to the same network as all the other methods. Other parts of the model remain the same.
We use the orthogonal flow. A lower learning rate of 0.0005 is used for stability.

For SIVI, we find the model is unstable for learning rate of 0.001, so we change it to 0.0001. 
It also takes more epochs to produce good samples, so we train for 1000 epochs. 
We use $J=10$ proposals from the Gaussian posterior.

For RWS, each parameter update is accompanied with
both wake and sleep updates of the encoder parameters, using $K=50$ 
proposals. A larger $K$ can cause lower signal-to-noise ratio of the update for 
the encoder network.

\begin{figure}[ht]
    \centering
    \includegraphics[height=0.42\textheight]{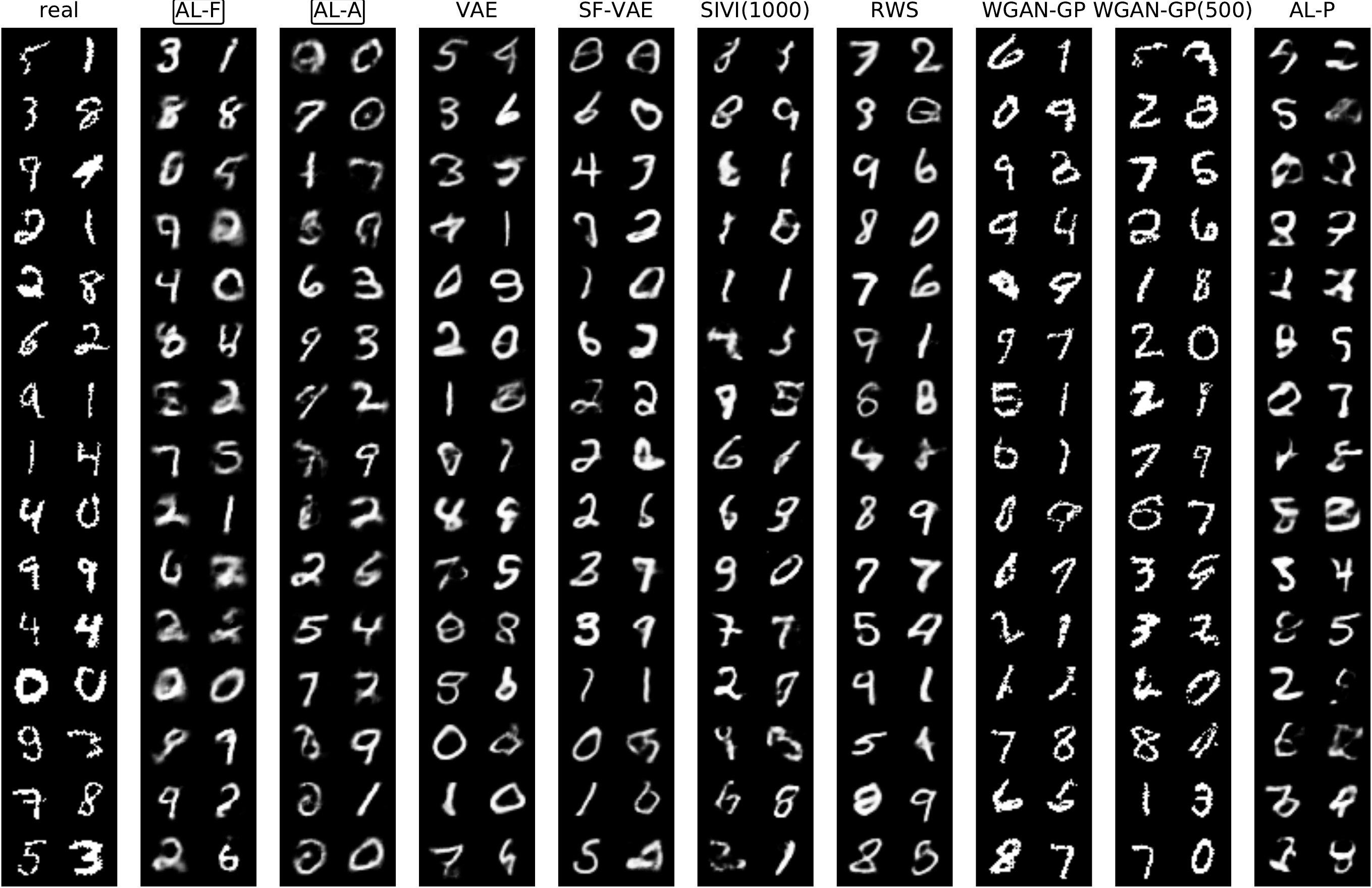}
    \caption{Samples for B-MNIST. Our main algorithms presented in the main text are highlighted in box. 
    Each model is trained for 50 epochs, except otherwise indicated in parenthesis next to 
    algorithm name.}
    \label{fig:real_mnist_0_conv}
\end{figure}

\begin{figure}[h!]
    \centering
    \includegraphics[height=0.42\textheight]{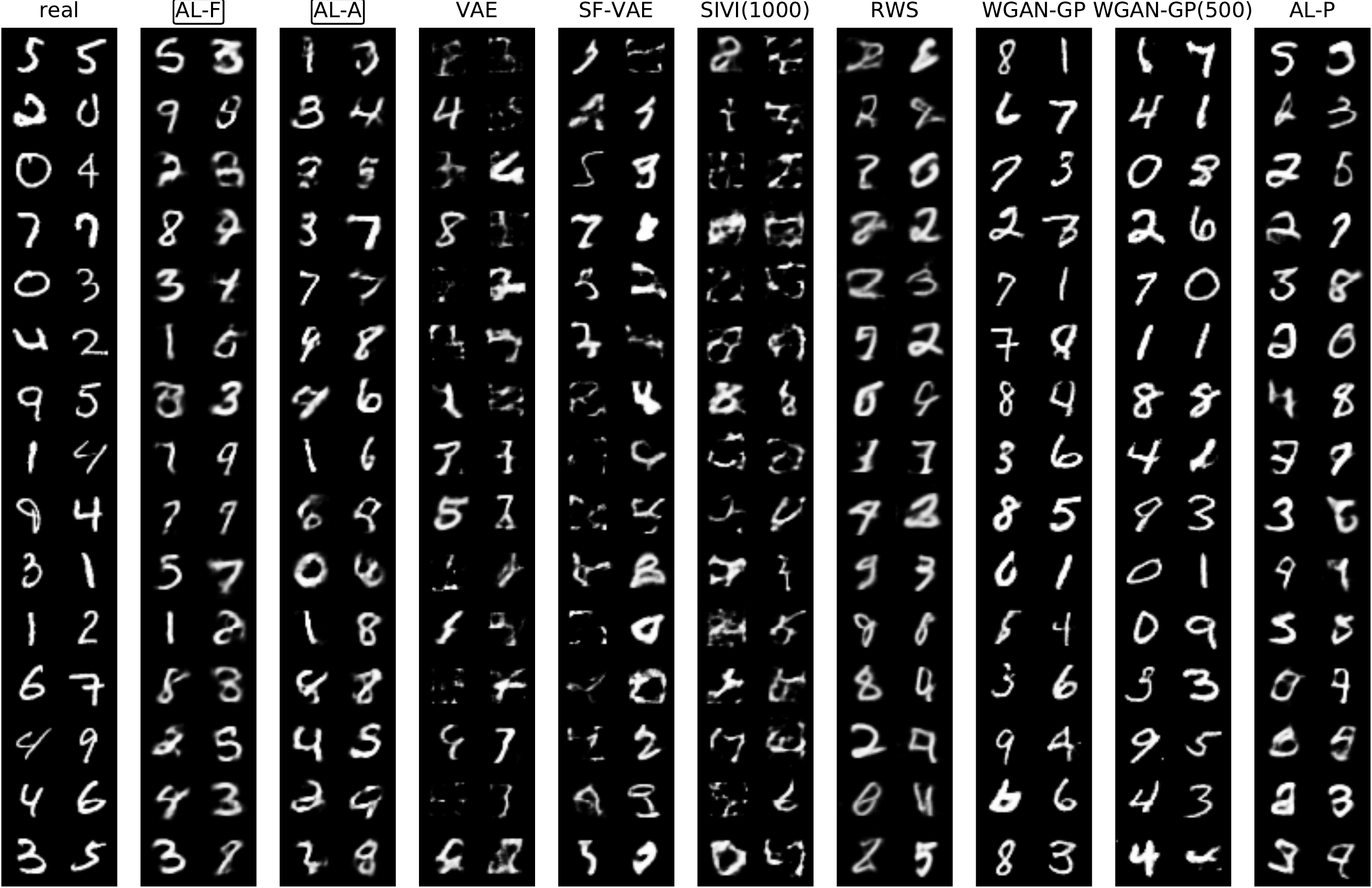}
    \caption{Samples for MNIST. Our main algorithm is highlighted in box. 
    Each model is trained for 50 epochs, except otherwise indicated in parenthesis next to 
    algorithm name.}
    \label{fig:real_mnist_1_conv}
\end{figure}

\begin{figure}[h!]
    \centering
    \includegraphics[height=0.42\textheight]{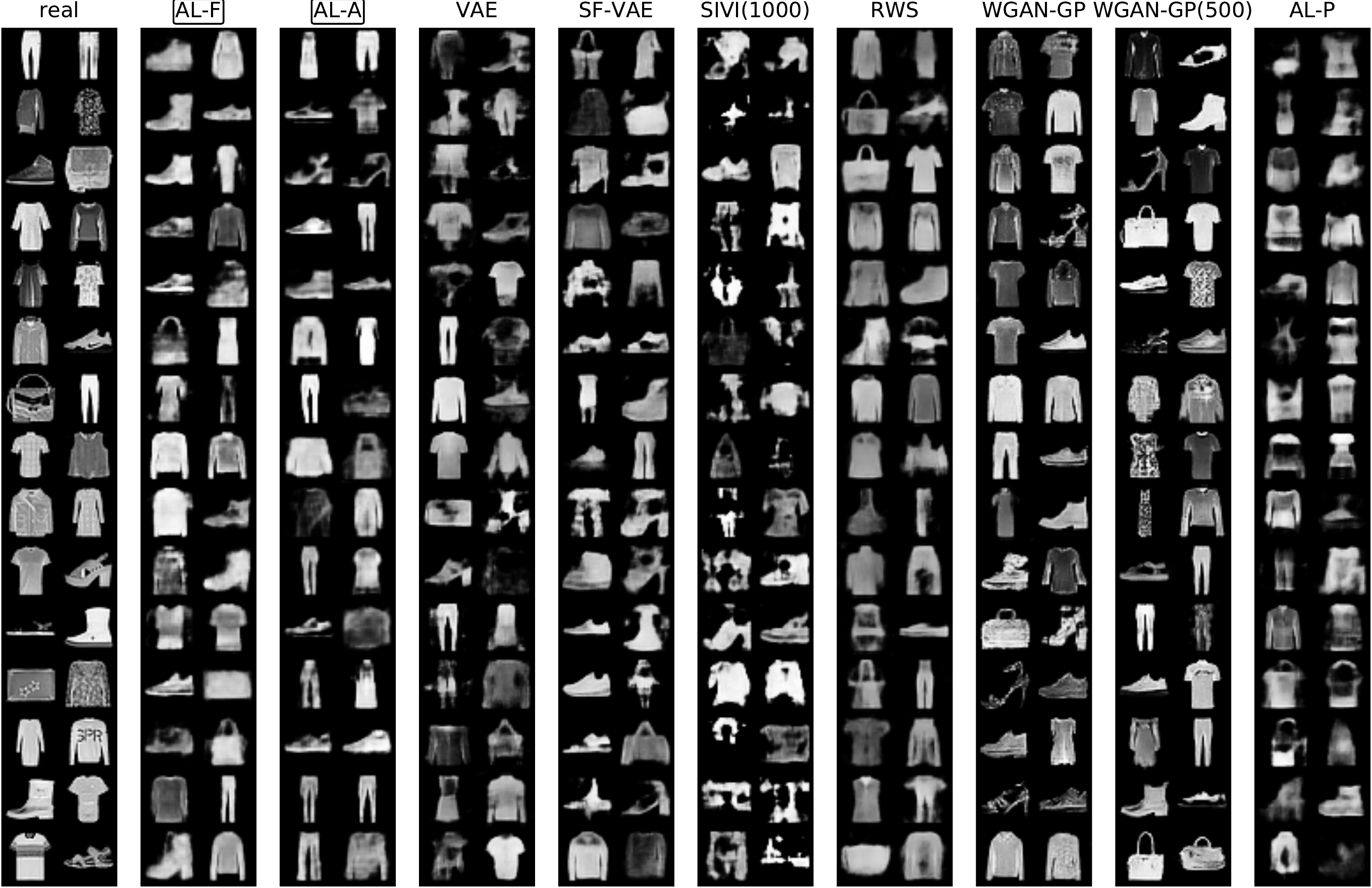}
    \caption{Samples for Fashion. Our main algorithms presented in the main text are highlighted in box. 
    Each model is trained for 50 epochs, except otherwise indicated in parenthesis next to 
    algorithm name.}
    \label{fig:real_fmnist_0_conv}
\end{figure}

\begin{figure}[h!]
    \centering
    \includegraphics[height=0.42\textheight]{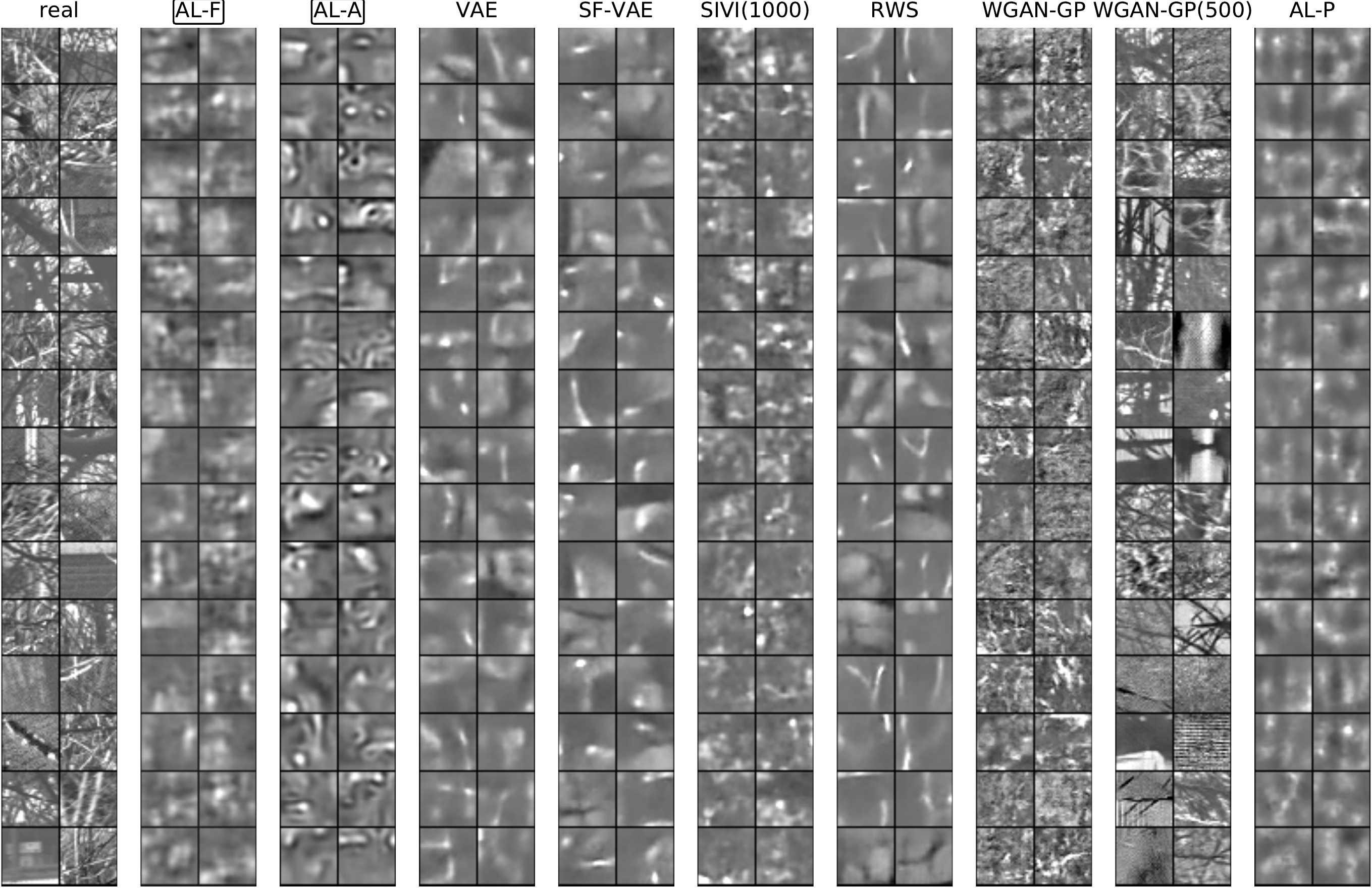}
    \caption{Samples for Natural. Our main algorithm is highlighted in box. 
    Each model is trained for 50 epochs, except otherwise indicated in parenthesis next to 
    algorithm name.}
    \label{fig:real_van_0_conv}
\end{figure}

\begin{figure}[h!]
    \centering
    \includegraphics[height=0.42\textheight]{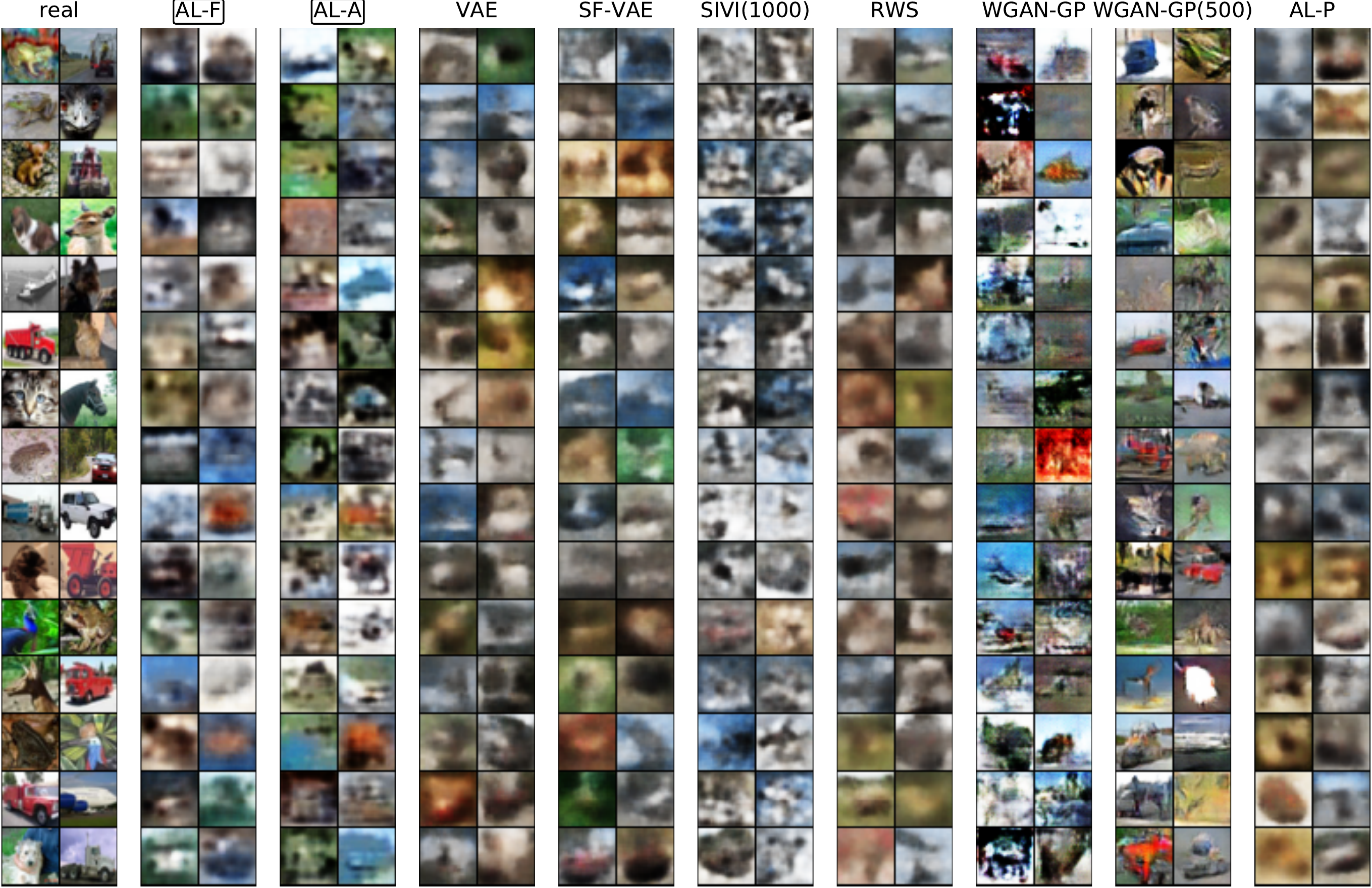}
    \caption{Samples for CIFAR-10. Our main algorithms presented in the main text are highlighted in box. 
    Each model is trained for 50 epochs, except otherwise indicated in parenthesis next to 
    algorithm name.}
    \label{fig:real_cifar_0_conv}
\end{figure}

\begin{figure}[h!]
    \centering
    \includegraphics[height=0.42\textheight]{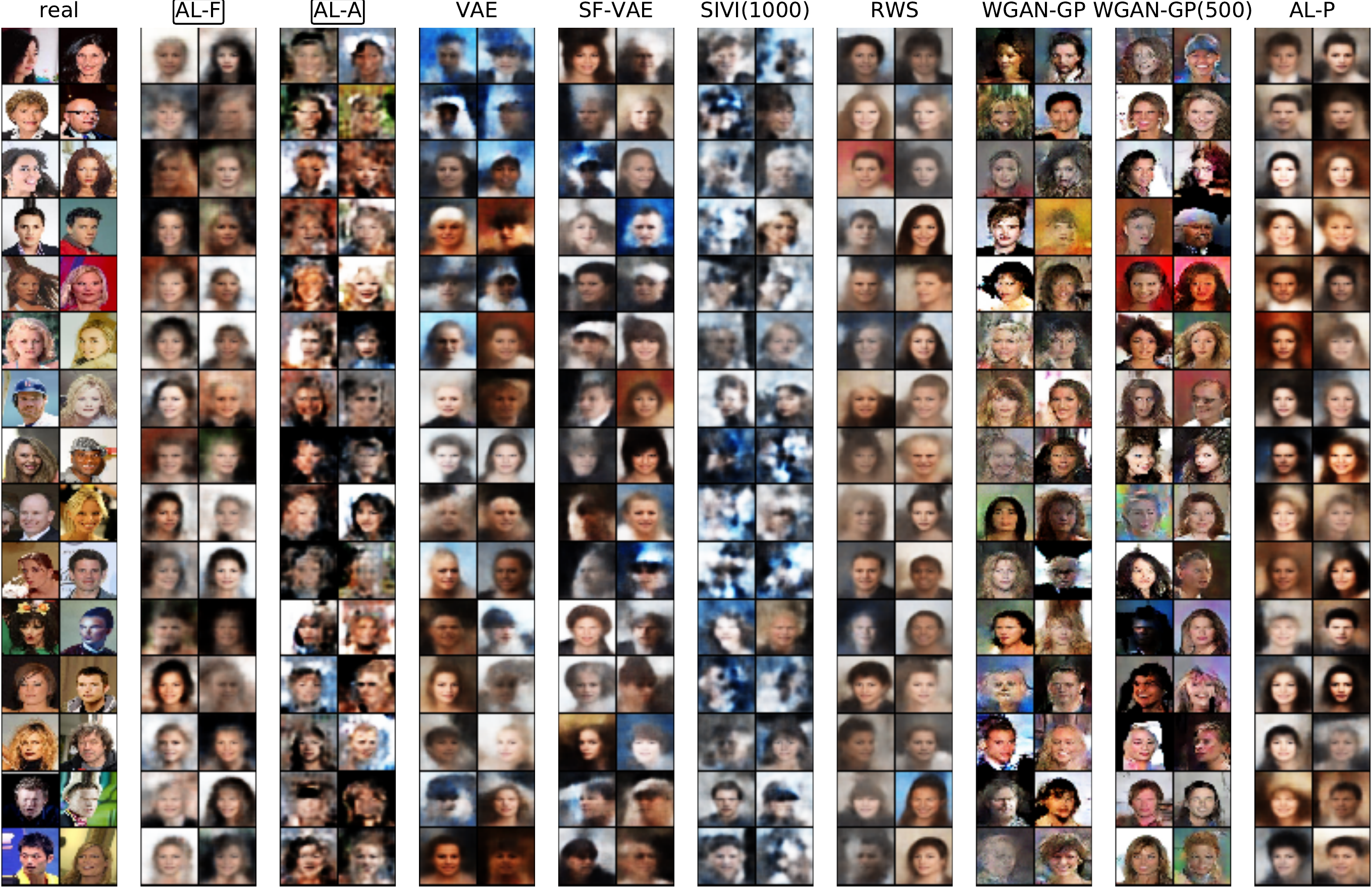}
    \caption{Samples for CelebA. Our main algorithms presented in the main text are highlighted in box. 
    Each model is trained for 50 epochs, except otherwise indicated in parenthesis next to 
    algorithm name.}
    \label{fig:real_celeb_0_conv}
\end{figure}

For WGAN-GP, learning is unstable for a learning rate of 0.001, so we train the model using a learning rate of 0.0001 for 50 epochs, which was not sufficient 
for it to produce good images. We also run WGAN-GP for 500 epochs 
on all datasets and show the samples from \cref{fig:real_mnist_0_conv} to \cref{fig:real_celeb_0_conv}.
We show the results of WGAN-GP just for reference, as
it is not trained using the maximum likelihood objective. 

To evaluate the quality, we use standard metrics FID and KID, which 
are computed using features of penultimate layers of neural networks 
pre-trained on relevant datasets. 
For both MNISTs, the features are from the LeNet trained to classify MNIST
digits. 
For Fashion, we used the LeNet network trained to classify the objects.
For Natural, CIFAR-10 and CelebA, we use inception network trained on 
ImageNet classification. For Natural, we duplicate the image along the channel axis to fill the three colour channels.

ALWS-A has lower FID and KID 
than other maximum likelihood methods in most cases, especially on original MNIST and Fashion MNIST,  but does not reach the level of WGAN-GP.

The KID and FID values during training are shown in \cref{fig:real_kid_epoch}. ALWS performs 
consistently better at every training epoch on all datasets except B-MNIST. 
On MNIST, ALWS-A converged the fastest and generates samples with stable quality. 
On CIFAR, ALWS-A and RWS converged faster than the others, but
VAE and Syl-VAE converge very slowly. We note that these figures are plotted  
against epochs, not wall-clock time. The run time
of ALWS is much longer than the other methods, taking around 3.5 seconds per iteration 
on a GeForce 1080 GPU, or 2.5 seconds on a Quadro P5000 with kernel adaptation. Nonetheless, this cost is worth the improvement over other maximum likelihood methods.

The samples from all methods are shown in \cref{fig:real_mnist_0_conv} to \cref{fig:real_celeb_0_conv}. These include 
samples from models presented in the main text, the WGAN-GP for 500 epochs, and the 
AL-P algorithm introduced in \cref{sec:particle_details}.

\clearpage


\begin{figure}[t]
    \centering
    \includegraphics[width=\textwidth]{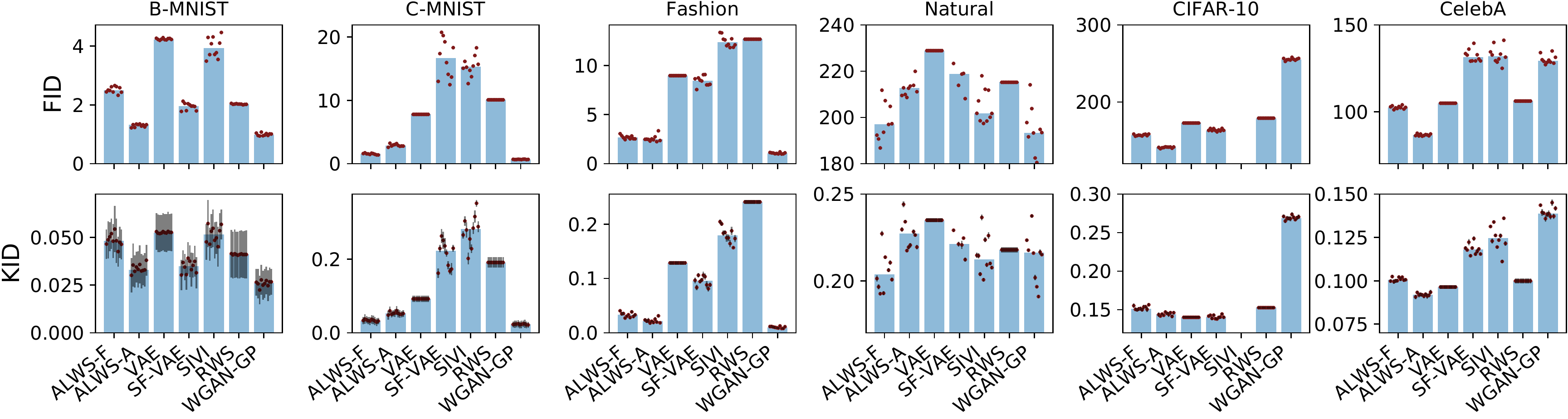}
    \caption{Same as \cref{fig:real_kid_fc} but using fully connected networks. None of the SIVI runs on CIFAR-10 converge.}
    \label{fig:real_kid_fc}
\end{figure}
\begin{figure}[h!]
    \centering
    \includegraphics[width=0.8\textwidth, trim=0 0 1.5cm 0, clip]{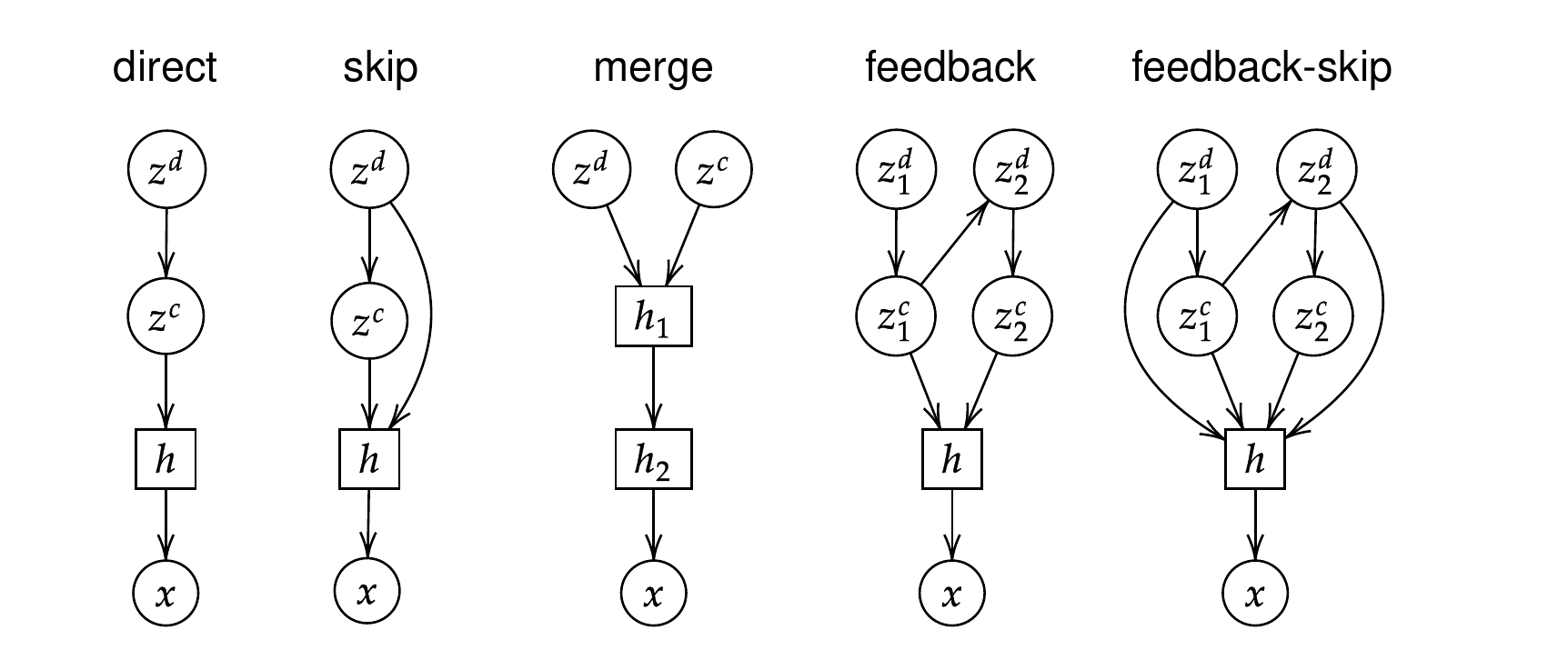}
    \includegraphics[width=\textwidth]{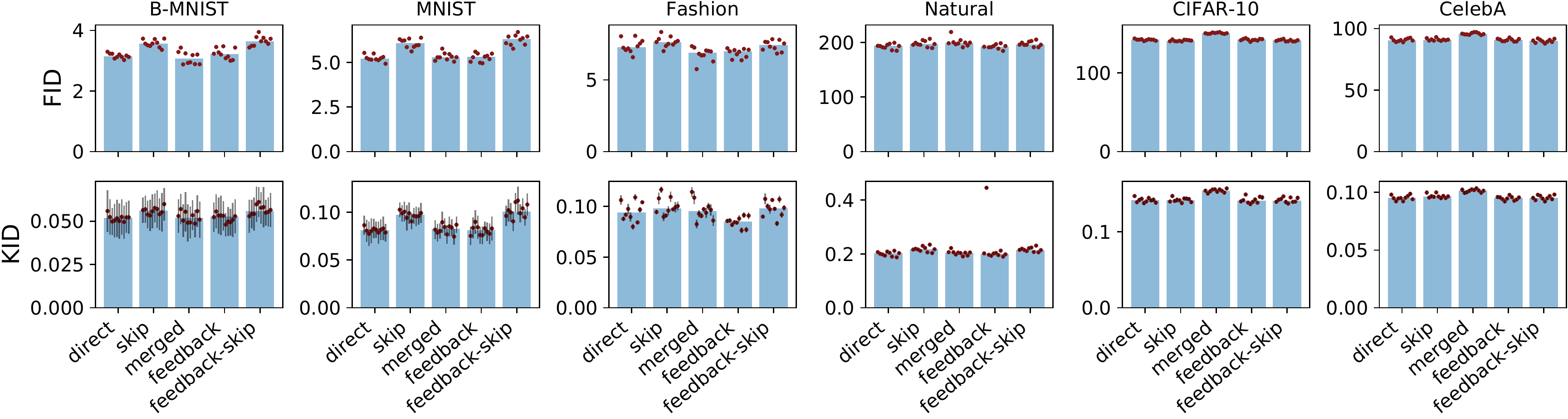}
    \caption{
    Top, the graphical representations of the generative models. 
    Circles indicate random variables, with $\vz^d$ as discrete Bernoulli and $\vz^c$ as  continuous Gaussian.
    Squares indicate deterministic nodes that are $\operatorname{ReLU}$ neurons activated 
    by nodes with incoming arrows. 
    The dimensionality of $\vz^d$ is 10, $\vz^c$ is 16, $\vz^d_1$ and $\vz^d_2$ are 5, and 
    $\vz^c_1$ and $\vz^c_2$ are 8.
    The node $h$ has 512 neurons.
    Bottom,
    FID and KID scores of generated images from architecturally complex models}
    \label{fig:real_kid_arch}
\end{figure}

\subsubsection{Results on fully connected networks}

We repeat the experiments for fully connected layers, with architecture $16\to512\to512\to$ image dimension.
The results are shown in \cref{fig:real_kid_fc}.
According to FID, models trained by ALWS out-perform other ML methods on all datasets except Natural.
KID agrees with FID except on CIFAR-10 where KID values are roughly the same for all ML methods.

\subsubsection{Results on complex generative networks} \label{sec:real_complex_details}

The goal here is to test how model architecture affects the quality of the generated samples.
Discrete variables can be used to capture features such as object category, 
so including these in the generative model may be beneficial. 
In order to train models with discrete latent variables, explicit reparameterisation schemes
have been developed in the past by continuous relaxation or overlapping transformation
\citep{JangPoole2017Categorical,VahdatAndriyash2018DVAE++,RolfeRolfe2017Discrete}, 
and has shown differential performances.
On the other hand,
amortised learning is agnostic to the discrete or continuous nature of the latents. 

We set out to explore different architectures while fixing the number of 
Bernoulli and Gaussian latent variables, respectively, and keep the number of parameters roughly the same.
The different graphs are depicted in \cref{fig:real_kid_arch} (top) and described in the legend.
The \textbf{direct} model is a simple chain graph. The top Bernoulli layer connects to a Gaussian layer, where the mean is a 
function of the Bernoulli, and the variance is fixed at 1.0. 
The \textbf{skip} model is similar to the direct model, except that it adds an 
additional connection from the discrete latents to the hidden units in the network.
The \textbf{merged} model combines the Bernoulli and Gaussian latents at the top layer, 
which goes through a first hidden $h_1$ layer of 16 units 
before feeding into the wide $h_2$ layer.
The \textbf{feedback} model has an architecture inspired by \citep{VahdatAndriyash2018DVAE++}. 
The latent $\vz^c_1$ parametrises the logits for $\vz^d_2$. 
The \textbf{feedback-skip} model is based on \textbf{feedback} and adds a skip connection to $h$ from the top Bernoulli layer.

The results are shown in \cref{fig:real_kid_arch} (bottom). 
Interestingly, we did not find any strong effect of model architecture on FID or KID. 
But the direct, merged and feedback architectures are clearly better than the other two
for the two MNIST datasets according to FID.

\end{document}

%% file: math_commands.tex
\usepackage{amsmath,amsfonts,bm}

\def\1{\bm{1}}

\def\vmu{{\bm{\mu}}}
\def\valpha{{\bm{\alpha}}}
\def\vomega{{\bm{\omega}}}
\def\vgamma{{\bm{\gamma}}}
\def\vrho{{\bm{\rho}}}

\def\vtheta{{\bm{\theta}}}
\def\vphi{{\bm{\phi}}}
\def\vsigma{{\bm{\sigma}}}
\def\veta{{\bm{\eta}}}

\def\vpsi{{\bm{\psi}}}
\def\vepsilon{\bm{\epsilon}}
\def\va{{\bm{a}}}
\def\vb{{\bm{b}}}

\def\ve{{\bm{e}}}
\def\vf{{\bm{f}}}

\def\vh{{\bm{h}}}

\def\vk{{\bm{k}}}
\def\vl{{\bm{l}}}
\def\vm{{\bm{m}}}
\def\vn{{\bm{n}}}
\def\vo{{\bm{o}}}

\def\vq{{\bm{q}}}
\def\vr{{\bm{r}}}
\def\vs{{\bm{s}}}

\def\vu{{\bm{u}}}
\def\vv{{\bm{v}}}
\def\vw{{\bm{w}}}
\def\vx{{\bm{x}}}
\def\vy{{\bm{y}}}
\def\vz{{\bm{z}}}

\def\evk{{k}}

\def\evz{{z}}

\def\mI{{\bm{I}}}

\def\mK{{\bm{K}}}

\def\mR{{\bm{R}}}

\def\mW{{\bm{W}}}

\def\mSigma{{\bm{\Sigma}}}

\DeclareMathAlphabet{\mathsfit}{\encodingdefault}{\sfdefault}{m}{sl}
\SetMathAlphabet{\mathsfit}{bold}{\encodingdefault}{\sfdefault}{bx}{n}

\def\gC{{\mathcal{C}}}
\def\gD{{\mathcal{D}}}
\def\gE{{\mathcal{E}}}
\def\gF{{\mathcal{F}}}

\def\gH{{\mathcal{H}}}

\def\gL{{\mathcal{L}}}

\def\gN{{\mathcal{N}}}

\def\gP{{\mathcal{P}}}

\def\gT{{\mathcal{T}}}
\def\gU{{\mathcal{U}}}

\def\gX{{\mathcal{X}}}
\def\gY{{\mathcal{Y}}}

\def\sH{{\mathbb{H}}}

\def\sR{{\mathbb{R}}}

\def\emK{{K}}

\DeclareMathOperator{\E}{\mathbb{E}}

\DeclareMathOperator{\Img}{\text{\Img}}
\DeclareMathOperator{\Rot}{\text{\Rot}}

\newcommand{\softplus}{\mathrm{softplus}}
\newcommand{\KL}{D_{\mathrm{KL}}}

\DeclareMathOperator*{\argmin}{arg\,min}

\newcommand{\ud}{\text{d}}